\documentclass[11pt]{article}

\usepackage[final]{acl}
\makeatletter
\newif\ifshowconferenceendmatter
\@ifpackagewith{acl}{preprint}{\showconferenceendmatterfalse}{\showconferenceendmattertrue}
\ifacl@anonymize
\else
\fi
\makeatother

\usepackage{times}
\usepackage{latexsym}

\usepackage[T1]{fontenc}
\usepackage[utf8]{inputenc}

\usepackage{microtype}

\usepackage{inconsolata}

\usepackage{graphicx}
\usepackage{fontawesome5}
\usepackage{eso-pic}
\usepackage{booktabs}
\usepackage{amsmath}
\usepackage{array}
\usepackage{xcolor}
\usepackage{colortbl}
\usepackage{tcolorbox}
\tcbuselibrary{breakable,skins}
\definecolor{promptborder}{HTML}{6686A6}
\definecolor{promptbg}{HTML}{F4F7FB}
\newtcolorbox{taskprompt}{enhanced,breakable,
  colback=promptbg,colframe=promptborder,
  colbacktitle=promptbg,coltitle=promptborder,
  title={Task Prompt},title after break={Task Prompt (continued)},
  fonttitle=\small\bfseries,fontupper=\small,
  boxrule=0.5pt,arc=2mm,
  left=7pt,right=7pt,top=5pt,bottom=6pt,
  before skip=8pt,after skip=10pt}
\definecolor{envborder}{HTML}{579187}
\definecolor{envbg}{HTML}{F3F8F6}
\newtcolorbox{taskenv}{enhanced,breakable,
  colback=envbg,colframe=envborder,
  colbacktitle=envbg,coltitle=envborder,
  title={Environment Structure},
  title after break={Environment Structure (continued)},
  fonttitle=\small\bfseries,
  fontupper=\fontsize{7}{8.5}\selectfont,
  boxrule=0.5pt,arc=2mm,
  left=7pt,right=7pt,top=5pt,bottom=6pt,
  before skip=8pt,after skip=8pt}
\newcommand{\envdir}[1]{\par\noindent\textcolor{envborder}{\bfseries\texttt{\detokenize{#1}/}}\par}
\newcommand{\envfile}[2]{\noindent\makebox[9pt][l]{\textcolor{envborder}{\rule{0.4pt}{5pt}\rule[2pt]{4pt}{0.4pt}}}\texttt{\detokenize{#1}}\hfill\textcolor{sdgray}{#2}\par}
\definecolor{sdgray}{gray}{0.55}
\newcommand{\sd}[1]{\textcolor{sdgray}{\scriptsize(#1)}}
\definecolor{gainfg}{rgb}{0.05,0.50,0.18}
\definecolor{gainbg}{rgb}{0.87,0.95,0.88}
\definecolor{rowblue}{rgb}{0.92,0.95,0.99}
\definecolor{lossfg}{rgb}{0.65,0.12,0.12}
\definecolor{lossbg}{rgb}{0.98,0.89,0.89}
\newcommand{\gain}[1]{\begingroup\setlength{\fboxsep}{1.2pt}\ifdim #1pt<0pt\colorbox{lossbg}{\textcolor{lossfg}{\scriptsize(#1)}}\else\colorbox{gainbg}{\textcolor{gainfg}{\scriptsize(+#1)}}\fi\endgroup}
\newcommand{\scorecell}[3]{\begin{tabular}{@{}c@{}}$#1$\,\sd{#2}\\[0.5pt]\gain{#3}\end{tabular}}
\newcommand{\modellogo}[1]{\makebox[1.4em][l]{\raisebox{-0.15ex}{\includegraphics[width=1.1em,height=1em,keepaspectratio]{assets/model_logos/#1.png}}}}
\newcommand{\paperresource}[3]{\href{#1}{\textcolor{envborder}{#2\hspace{0.35em}#3}}}

\title{Data World: Scaling Data Environments from Scratch}

\author{
  \textbf{Zhanli Li\textsuperscript{3,4}$^{\dagger,\S}$},
  \textbf{Zeyuan Xu\textsuperscript{5}},
  \textbf{Chaoxu Pang\textsuperscript{3}},
  \textbf{Zeyuan Liu\textsuperscript{3}},
  \textbf{Anyang Ji\textsuperscript{1,2,3}$^{\dagger}$},
  \textbf{Te Sun\textsuperscript{6}},
\\
  \textbf{Kongcheng Zhang\textsuperscript{3,7}$^{\dagger}$},
  \textbf{Wenbo Xu\textsuperscript{3,8}$^{\dagger}$},
  \textbf{Xubin Yue\textsuperscript{3,7}$^{\dagger}$},
  \textbf{Shipeng Luo\textsuperscript{9}},
\\
  \textbf{Junjie Zhao\textsuperscript{6}},
  \textbf{Yingfan Hua\textsuperscript{10}},
  \textbf{Jingqing Ruan\textsuperscript{3}},
  \textbf{Zhongda Su\textsuperscript{3}},
  \textbf{Feng Hong\textsuperscript{3}$^{\ast}$},
\\
  \textbf{Zhizhao Zeng\textsuperscript{3}},
  \textbf{Ke Zeng\textsuperscript{3}},
  \textbf{Xunliang Cai\textsuperscript{3}},
  \textbf{Hanjia Ye\textsuperscript{1,2}$^{\ast}$}
\\[6pt]
  \footnotesize
  \textsuperscript{1}School of Artificial Intelligence and
  \textsuperscript{2}National Key Laboratory for Novel Software Technology,
  Nanjing University\\
  \footnotesize
  \textsuperscript{3}Longcat Interaction Team, Meituan,
  \textsuperscript{4}Zhongnan University of Economics and Law\\
  \footnotesize
  \textsuperscript{5}Hunan University,
  \textsuperscript{6}Shanghai Jiao Tong University,
  \textsuperscript{7}Zhejiang University,
  \textsuperscript{8}Nanjing University\\
  \footnotesize
  \textsuperscript{9}Institute of Automation, Chinese Academy of Sciences,
  \textsuperscript{10}The Chinese University of Hong Kong
\\[4pt]
  \small
  $^{\ast}$Correspondence:
  \href{mailto:hongfeng03@meituan.com}{hongfeng03@meituan.com},
  \href{mailto:yehj@lamda.nju.edu.cn}{yehj@lamda.nju.edu.cn}\\[2pt]
  \footnotesize
  $^{\S}$First author:
  \href{mailto:lizhanli@stu.zuel.edu.cn}{lizhanli@stu.zuel.edu.cn}.
  $^{\dagger}$Work done during an internship at Longcat Interaction Team, Meituan.
\\[5pt]
  \small\normalfont
  \paperresource{https://github.com/Zhanli-Li/Data-World}{\faIcon{github}}{GitHub}
  \quad\textcolor{sdgray}{\textbar}\quad
  \paperresource{https://huggingface.co/collections/Zhanli-Li/dataworld}{\raisebox{-0.2ex}{\includegraphics[height=1.3em]{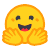}}}{Hugging Face}
  \quad\textcolor{sdgray}{\textbar}\quad
  \paperresource{https://zhanli-li.github.io/Data-World/}{\faIcon{globe}}{Project Page}
}

\begin{document}
\makeatletter
\ifacl@anonymize\else
  \AddToShipoutPictureFG*{%
    \AtPageUpperLeft{%
      \put(\LenToUnit{15mm},\LenToUnit{-7mm}){%
        \raisebox{-\height}{\includegraphics[width=42mm]{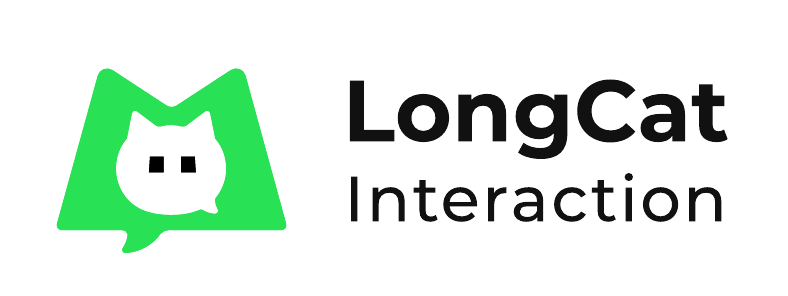}}%
      }%
      \put(\LenToUnit{58mm},\LenToUnit{-11.2mm}){%
        \raisebox{-\height}{\includegraphics[width=42mm]{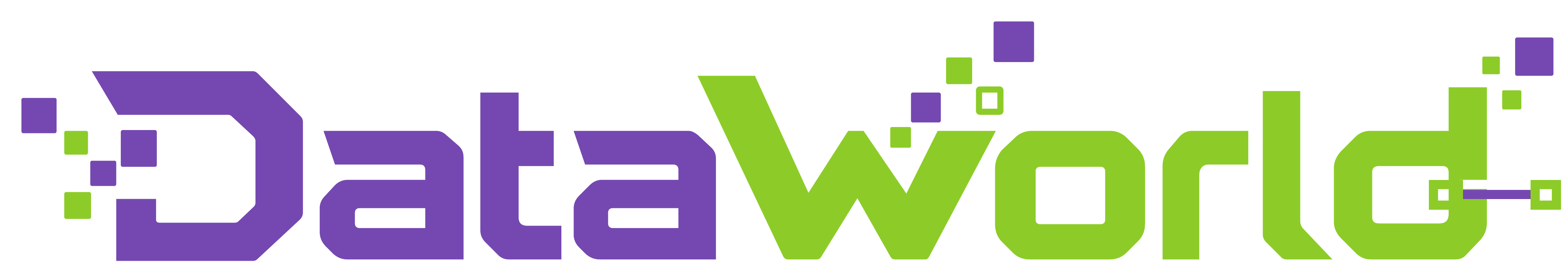}}%
      }%
    }%
  }
\fi
\makeatother
\maketitle

\begin{abstract}
Data agents use code to process heterogeneous files and derive analytical insights. Scaling their training faces costly environment curation, privacy and intellectual-property restrictions, and difficulty verifying open-ended findings without known generative truth. Our core insight is to specify the world behind the files before generating the files themselves. We introduce \textbf{Data World}, a multi-agent engine that synthesizes environments, tasks, and rubrics from explicit mechanisms. Independent audits check target recoverability from observed files; rubrics support evaluation and RL rewards. Human preference-based curation for question realism selects 100 tasks across 50 held-out worlds and 16 domains for \textbf{DataWorld-Bench}. The engine also yields 2{,}135 training tasks from 1{,}037 disjoint worlds. Evaluation of ten frontier models shows that the benchmark remains challenging, particularly for insight discovery, with a consistent computation--insight gap. RL-zero training on the synthesized tasks produces \textbf{DataWorld-35B-A3B} from Qwen3.6-35B-A3B without trajectory distillation. Under matched evaluation, it improves over its base by $7.06$ points on DataWorld-Bench, $14.28$ on DAComp, $20.90$ on DABstep-Hard, and $12.13$ on JobBench, a broader occupational benchmark.
\end{abstract}

\section{Introduction}
\begin{figure}[t]
  \centering
  \includegraphics[width=\columnwidth,trim=6.64bp 11.68bp 505.12bp 0bp,clip]{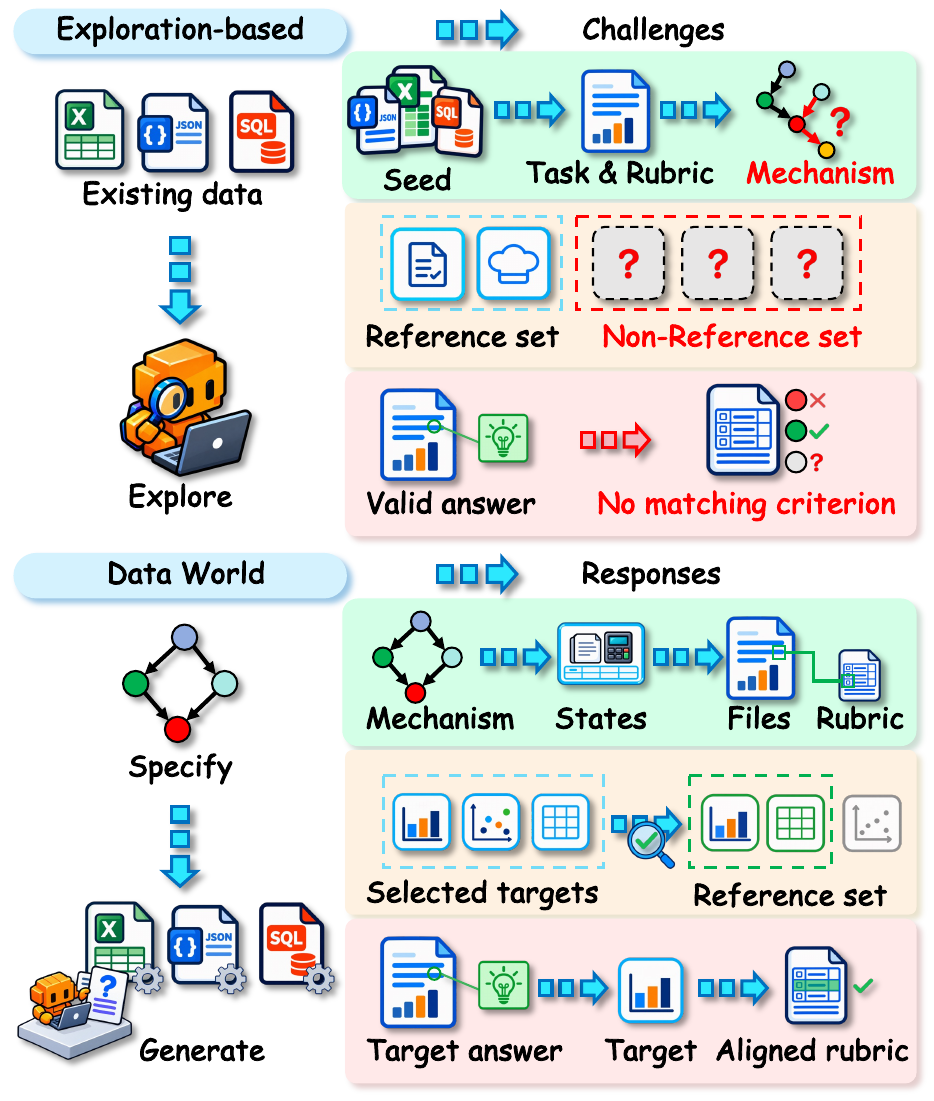}
  \caption{Supervision for data agents: exploration-based synthesis (top) versus Data World (bottom). Exploring existing files may leave valid findings without matching criteria; Data World specifies mechanisms first, aligning files and rubrics with analytical targets.}
  \label{fig:exploration-comparison}
\end{figure}
Data agents use code to inspect heterogeneous files, compute quantities, and derive evidence-supported analytical insights \citep{huang2024dacode,hong2025datainterpreter}. Their \emph{data environments} often contain multi-source structured data distributed across CSV and Excel tables \citep{dong2026finch,hasegawa2026datagovbench}. A task may ask an agent to explain a change in an outcome or trace its downstream implications, requiring more than correct computation alone. Training and evaluating these agents therefore require both diverse environments for multi-step analysis and feedback that distinguishes supported conclusions from errors. As agentic LLM post-training scales, this joint supply of environments and supervision becomes a central constraint \citep{huang2025envscaling,wang2026awm,tu2026scaleenv,tian2026astra,xu2026envfactory}.

Synthetic data has long served as an important complement to real data when the latter is costly or restricted \citep{walonoski2018synthea}. Yet for data agents, two obstacles limit this supply. First, \emph{diversity and accessibility}: acquiring and annotating real workflows is costly, while privacy and intellectual-property restrictions constrain reuse \citep{dong2026finch,walonoski2018synthea}. Second, \emph{verifiability}: the data-generating process behind real files is often unavailable, so open-ended findings rely on curated references or approximate matching \citep{gu2024blade,sahu2024insightbench,zhang2025datascibench}. Exploring these files to write tasks and rubrics ties supervision to what the curator or mining model discovers \citep{ramrakhya2025taskgen,liu2026datasage}. For example, consider an expert or strong model given food-delivery records collected from an online platform and asked to write a rubric for evaluating analyses of declining sales. Inspecting these records, it may identify expired promotions as a contributor but overlook reduced restaurant opening hours. An agent uncovering the latter may then receive no credit because the rubric omits that explanation, even when the records support it. Indeed, ten experts or strong models could offer ten different explanations for the same decline, as data complexity and inadequately controlled confounders can leave the underlying mechanism ambiguous. How can we scale diverse data environments with grounded supervision while retaining open-ended analytical depth?

Our core insight is to specify the world behind the files before generating the files themselves. Analysis depends on statistical and relational structure that explicit mechanisms can instantiate in code. This enables varied environments without private source records, while known generative states ground checks of selected analytical targets. Synthesizing new environments and task-specific answers also reduces the risk of pretraining contamination associated with reusing public datasets. Figure~\ref{fig:exploration-comparison} illustrates the resulting difference for a data agent: its targeted findings have aligned rubric criteria rather than depending on prior exploration of existing files. The key challenge is preserving target recoverability through imperfect observations without making the analysis trivial.

We propose \textbf{Data World}, a multi-agent synthesis engine that expands a directed acyclic dependency graph (DAG) into a world contract, generates a canonical state layer by layer, and projects it into analyst-visible files. Independent agent sessions exchange read-only artifacts to preserve cross-layer consistency; a separate auditor checks the specified relationships and target-phenomenon recoverability from imperfect observations. The generative record then grounds tasks and rubrics, while the solving agent receives only the task and observed files; rubrics distinguish \emph{data computation} from \emph{insight discovery}, and scoring against them supplies evaluation feedback and training rewards. Running this engine yields \textbf{DataWorld-Bench}, with 100 tasks from 50 held-out worlds across 16 domains, and a world-disjoint RL training set of 2{,}135 tasks from 1{,}037 worlds. Evaluation of ten frontier models reveals a consistent computation--insight gap, and RL-zero training with rubric rewards produces \textbf{DataWorld-35B-A3B}, which improves over its base on held-out worlds and three external benchmarks spanning data analysis and broader professional work.
Our contributions are summarized as follows:
\begin{itemize}
  \item We introduce a mechanism-first framework for synthesizing data environments, grounding task rubrics in a canonical generative state and auditing the recoverability of target phenomena from imperfect observations.
  \item We derive DataWorld-Bench and a world-disjoint training set from this engine, using fine-grained rubrics to reveal a computation--insight gap across ten frontier models.
  \item We demonstrate training utility via RL-zero training on the synthesized tasks, without distilling trajectories from stronger models, yielding substantial gains over the base on held-out worlds and three external benchmarks.
\end{itemize}

\section{Related Work}

\paragraph{Data analysis agents and benchmarks.}
Existing environments draw on Kaggle competitions, scientific datasets, government open data, and released enterprise records \citep{jing2024dsbench,gu2024blade,hasegawa2026datagovbench,dong2026finch}.
LLM data-agent benchmarks span text-to-SQL over curated databases \citep{yu2018spider,li2023bird}, code-driven analysis over files \citep{hu2024infiagent,jing2024dsbench,huang2024dacode,egg2025dabstep,zhang2025datascibench,liu2024qrdata,hong2025datainterpreter}, and multi-table government or enterprise spreadsheet workflows \citep{hasegawa2026datagovbench,dong2026finch}.
Closest to our setting, open-ended insight-discovery benchmarks---DiscoveryBench and BLADE for scientific discovery \citep{majumder2024discoverybench,gu2024blade}, InsightBench and InsightEval for business analytics \citep{sahu2024insightbench,zhu2025insighteval}---score the insights agents extract, but their environments come from available sources and their reference insights from expert annotation or model-based exploration.
Data World complements these source-based benchmarks with mechanism-grounded criteria and audits of target-phenomenon recoverability.

\paragraph{Task synthesis and rubric-based rewards.}
A complementary line of work keeps the data real but synthesizes \emph{tasks} by exploring it: agents probe an environment to propose diverse, feasible, and verifiable tasks \citep{ramrakhya2025taskgen}, or mine insights from available datasets to build evaluation suites \citep{sahu2024insightbench}.
When the underlying generative process is unavailable, scaling challenging open-ended tasks alongside dependable supervision requires additional curation or evaluation machinery \citep{gu2024blade,liu2026datasage}.
Rubric-derived rewards scored by LLM judges extend reinforcement learning to open-ended tasks \citep{gunjal2025rar,huang2025rubicon,wei2026qurl}. Our rubrics are instead written against a generative record with independent audit results: the judge still interprets responses, but specified criteria can be cross-checked against that record and the observed files. This limits dependence on an annotator's or mining model's ability to discover the targeted insight from existing data, without removing biases in mechanism design, rubric selection, or judging.

\paragraph{Synthetic environments and data.}
Environments are widely seen as the bottleneck of agent post-training \citep{huang2025envscaling}, yet early interactive environments such as WebArena \citep{zhou2023webarena} and $\tau$-bench \citep{yao2024taubench} are hand-curated and hard to scale.
Recent pipelines synthesize environments for MCP-based tool-use agents in different ways.
Agent World Model \citep{wang2026awm} and ScaleEnv \citep{tu2026scaleenv} grow simulated tools, databases, and tasks from platform names or domain keywords, without assuming API documentation or predefined tasks.
EnvScaler \citep{song2026envscaler} and ASTRA \citep{tian2026astra} distill environments from existing trajectories and task sets, while EnvFactory \citep{xu2026envfactory} and Agent-World \citep{dong2026agentworld} assemble environments from web-found MCP interfaces.
In contrast, Data World targets terminal-based analysis of interdependent files, with specified generative relationships for auditing analytical findings rather than simulated tool APIs.
Synthetic \emph{data} methods such as GReaT, SDV, and REaLTabFormer learn from existing databases \citep{borisov2023great,patki2016sdv,solatorio2023realtabformer}; Synthea simulates records from explicit mechanisms \citep{walonoski2018synthea}, and synthetic causal benchmarks target causal discovery \citep{ferdous2025timegraph}. Our focus is the recoverability of analytical targets across imperfect, interdependent files.

\section{Method}
\label{sec:method}

\begin{figure*}[t]
  \centering
  \includegraphics[width=\textwidth]{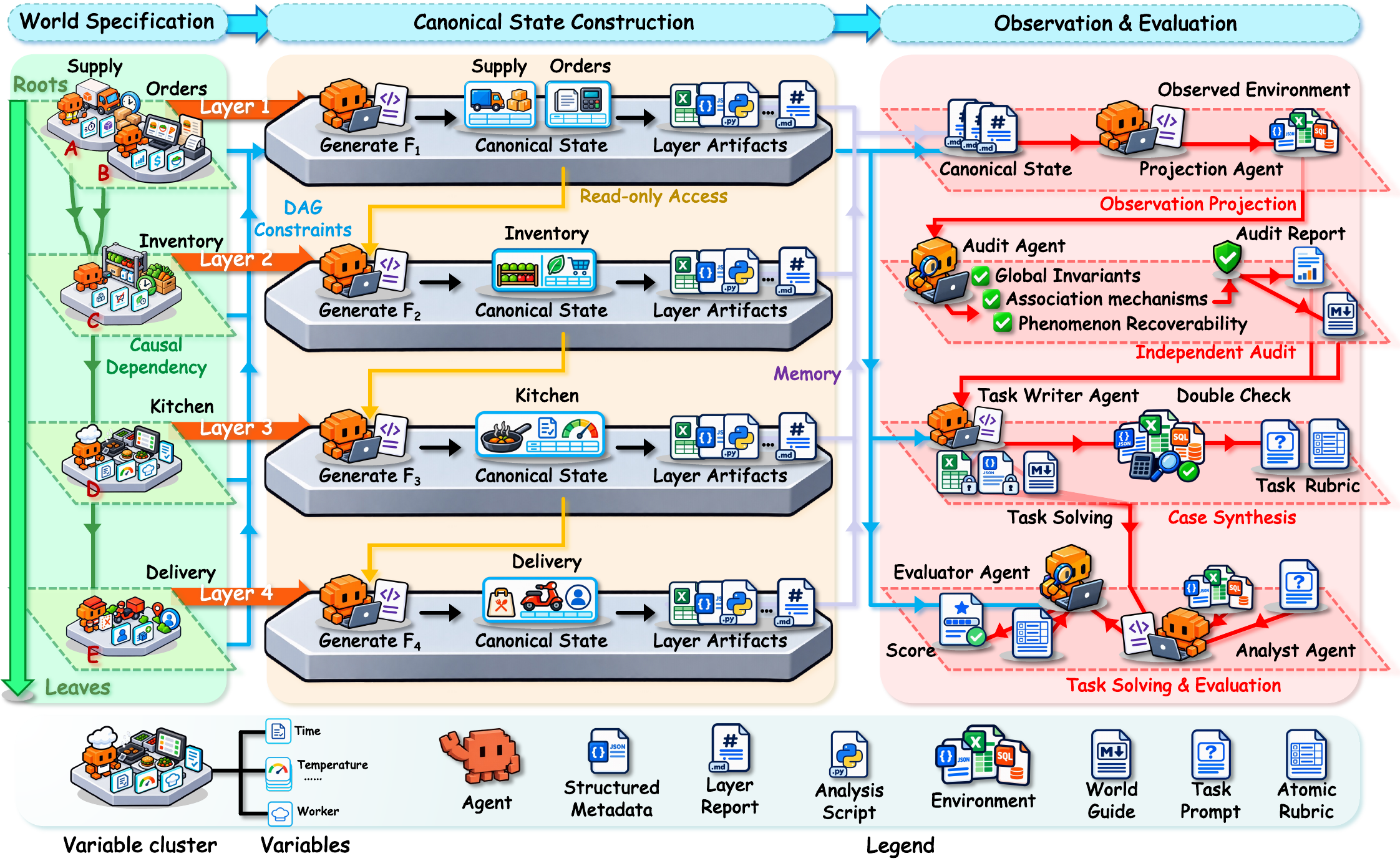}
  \caption{Data World synthesis and evaluation pipeline. A dependency DAG and world contract guide layer-wise canonical-state generation; observation projection produces analyst-visible files. An independent auditor checks relationships, invariants, and target recoverability. The generative record grounds tasks and rubrics; the data agent solves tasks from observed files, and a separate evaluator grades its report.}
  \label{fig:data-world-overview}
\end{figure*}

Data World generates environments and checkable analytical targets together by specifying the mechanisms behind the files first. Figure~\ref{fig:data-world-overview} summarizes the multi-agent pipeline, from world specification and generation to auditing, task synthesis, and agent evaluation.

\subsection{Formalization and Overview}

We use \textbf{Data World} to refer to our framework and formalize a synthesized \emph{world} as a four-tuple $\mathcal{W}=(\mathcal{G},\,\mathcal{X},\,\mathcal{E},\,\mathcal{I})$.
The \emph{dependency skeleton} $\mathcal{G}=(V,E)$ is a directed acyclic graph in which each node $v\in V$ represents a cluster of variables, and each edge $(u,v)\in E$ specifies a directional generative relationship (e.g., monotonic, threshold, saturation, lagged, or interactive), described by $\mathrm{note}(u,v)$. Edge directions encode generation order and modeled influences.
$\mathcal{X}$ is the \emph{canonical state}, i.e., the objective state of the world that governs the phenomena under analysis.
The \emph{data environment} $\mathcal{E}$ is the concrete directory of tables, exports, and logs visible to the analyst, produced by applying an observation map $O$ to $\mathcal{X}$.
$\mathcal{I}$ is a set of \emph{recomputable phenomena}: specified relationships or quantitative findings in the generated world that can be re-derived and checked from $\mathcal{E}$.
The pipeline first synthesizes and audits worlds, then generates tasks from them. A separate curation step selects a subset for benchmark evaluation:
\begin{equation}
\begin{gathered}
\mathcal{G}\longrightarrow\mathcal{C}\longrightarrow\mathcal{X}
\longrightarrow\mathcal{E}=O(\mathcal{X}) \\
\xrightarrow{\text{audit}}\mathcal{W}
\xrightarrow{\text{task synthesis}}\mathcal{T}.
\end{gathered}
\end{equation}
Here $\mathcal{C}$ is the \emph{world contract}, and $\mathcal{T}$ is the pool of $(\textit{task},\textit{rubric})$ pairs generated from a world $\mathcal{W}$ with its audit record. The test split requires all audit checks to pass; the training split uses a less restrictive filter.
Each stage uses an independent agent session; canonical-state generation assigns one session per DAG layer. Sessions share no conversational state and communicate only through artifacts on disk.
Each generation stage records a \emph{machine-checkable ledger} for verification and a \emph{narrative self-report} to convey its design decisions to later stages and the auditor.

\subsection{Environment Synthesis}

\paragraph{Dependency Skeleton.}
We use personas from PersonaHub\footnote{\url{https://huggingface.co/datasets/proj-persona/PersonaHub}} to prompt an LLM to construct synthetic dependency DAGs reflecting the contexts described by those personas. These DAGs serve as the skeletons $\mathcal{G}$ for subsequent world generation.

\paragraph{World Contract.}
Given $\mathcal{G}$, a contract-writing agent materializes it into a structured \emph{world contract} $\mathcal{C}$ that imposes global consistency constraints on the entire generation process.
The contract declares: the entities and timeline of the world; the \emph{generative mechanism} attached to each directed edge (functional form, heterogeneity, temporal behavior, confounders); a set of \emph{global invariants} (identifiers, units, and shared events that must remain consistent across layers); a set of \emph{observation systems} (the imperfect business systems through which the truth will be observed); and the final deliverables, including the datasets and the recomputable phenomena $\mathcal{I}$.
The contract is fixed before data generation and constrains all subsequent stages.

\paragraph{Canonical Truth Generation.}
Per-layer agents generate the canonical state in topological order along the dependency DAG $\mathcal{G}$.
Specifically, we compute a topological stratification $V=\bigsqcup_{\ell=1}^{L}V_\ell$ of $\mathcal{G}$ according to the longest-path depth $d(v)$ from source nodes ($d=0$), setting $V_\ell=\{\,v:d(v)=\ell-1\,\}$ so that every node is placed in a strictly later layer than all of its parents.
Each layer $\ell$ is instantiated by its own agent session:
\begin{equation}
\mathcal{X}_{V_\ell}=F_\ell\big(\mathcal{X}_{\mathrm{pa}(V_\ell)};\,\mathrm{seed}_\ell\big),
\end{equation}
where the upstream state $\mathcal{X}_{\mathrm{pa}(V_\ell)}$ produced by prior layer agents is strictly read-only, and $F_\ell$ is a deterministic script written and executed by the current layer agent under a fixed random seed.
The outputs are committed to disk as immutable inputs for subsequent layers.
This topology-driven generation uses two consistency safeguards:
(i) \textbf{Transactionality}: all nodes within a layer are generated jointly and committed atomically; no half-finished state may exist.
(ii) \textbf{Dependency checks}: for every incoming edge $(u,v)$, the layer agent runs empirical checks on the generated data for the specified association pattern, functional form, time lag, and heterogeneity declared in $\mathrm{note}(u,v)$, recording the results in its ledger before propagating to downstream layers.

\paragraph{Observation Projection.}
An independent session applies the observation map $O$ to the canonical state $\mathcal{X}$ to produce the analyst-visible files $\mathcal{E}$.
$O$ is not arbitrary noise, but a composition of \emph{explicit observation mechanisms}: the same entity carries different identifier spellings and key conventions across systems; defects such as missing values, delays, duplicates, unit drift, categorical encodings, and schema drift each correspond to a concrete observation process (who introduced it, at which step, and why); files are cross-referenced through unlabeled foreign keys, with no data dictionary provided.
Projection preserves the canonical state, but recovering findings from imperfect observations requires cross-system reconciliation. The auditor checks that each phenomenon in $\mathcal{I}$ remains recoverable from $\mathcal{E}$ without access to that state.
\paragraph{Independent Audit.}
Finally, an \emph{independent auditor} recomputes the completed world using its artifacts: the skeleton, the contract, the ledgers and self-reports, the generation scripts, the canonical state, and the observed environment.
The auditor writes its own scripts to re-check each specified edge relationship, global invariant, and phenomenon in $\mathcal{I}$, and additionally writes a world guide from the perspective of an analyst inside the world.
The auditor is forbidden from modifying the environment or the truth, and can only issue a pass/fail verdict; failed expectations must be faithfully reported.

\subsection{Task Synthesis}

Unlike prior work that explores fixed environments or curates existing files to propose tasks \citep{ramrakhya2025taskgen,sahu2024insightbench}, Data World grounds tasks directly in the audited generative process.
Each task is anchored to a single target node and asks the data agent either to explain its variation using upstream factors (the \emph{cause} route, following in-edges) or to trace its implications for downstream outcomes (the \emph{consequence} route, following out-edges). For example, an agent may investigate factors behind encounter volumes or trace production losses through inventories to downstream availability. These route names denote directions through the generative graph.
To ensure uniform coverage across the skeleton, we partition nodes into $K$ depth strata $S_k=\{\,v:\lfloor d(v)K/(d_{\max}+1)\rfloor=k-1\,\}$ using the longest-path depth $d(v)$, and deterministically select the maximal-degree nodes in each stratum:
\begin{equation}
\begin{split}
t^{\mathrm{cause}}_k&=\arg\max_{v\in S_k}\deg^{-}(v),\\
t^{\mathrm{cons}}_k&=\arg\max_{v\in S_k}\deg^{+}(v).
\end{split}
\end{equation}
This yields up to $2K$ structurally diverse tasks per world. Section~\ref{app:task-cases} presents environment descriptions, complete file structures, and corresponding task prompts for public safety, class-capacity planning, and production yield, illustrating the cross-system analysis required along the cause and consequence routes.

For each sampled target, a question-writer agent produces a $(\textit{task},\ \textit{rubric})$ pair using read-only access to the contract, generation scripts, ledgers and self-reports, canonical state, and world guide (\texttt{WORLD.md}). It checks that the targeted phenomena are derivable from $\mathcal{E}$ before writing criteria grounded in this generative provenance. Section~\ref{app:intuitive-theory} offers an intuitive analysis of this information advantage for rubric coverage.
Each rubric separates \emph{data computation} (processing and numerical estimates) from \emph{insight discovery} (conclusions supported by those estimates). Both sections contain atomic positive items for supported calculations or interpretations and negative items for demonstrably wrong calculations or contradicted conclusions. Valid insights outside the reference list can occur \citep{zhu2025insighteval}; they receive no credit but no penalty merely for being unlisted.
The evaluator has access to the audited truth and scores reports item by item, normalizing net rubric points to $[0,1]$.

\subsection{Benchmark and Training-Set Curation}
\label{sec:curation}

\begin{figure}[t]
  \centering
  \includegraphics[width=\columnwidth]{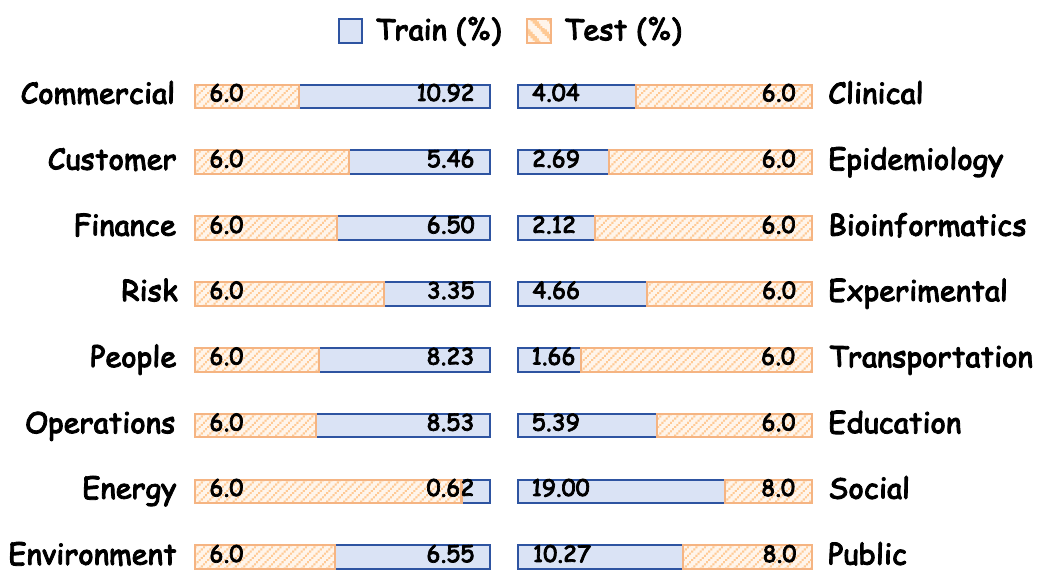}
  \caption{Taxonomy shares (\%) of the DataWorld-Bench test tasks and Data World RL training tasks.}
  \label{fig:split-distribution}
\end{figure}

Of 1{,}302 candidate worlds, we retain 1{,}087: 1{,}037 training worlds yielding 2{,}135 RL tasks and 50 held-out worlds yielding 100 test tasks across all 16 domains. Test worlds are entirely excluded from training and must pass every audit check; training worlds have a 95.98\% DAG-edge pass rate.
For each test world, we nominate one cause and one consequence task. A team of 15 annotators compares nominees with alternatives from the same world and route for question realism, with disagreements adjudicated. The selected tasks form DataWorld-Bench. Test tasks cover the 16 categories approximately uniformly, whereas training tasks are less evenly distributed (Figure~\ref{fig:split-distribution}).

\section{Experiments}

The mechanism-first design is useful for data agents only if the generated files support meaningful analysis and the associated rubrics provide useful learning signals. We therefore evaluate two questions: what computational and insight-discovery capabilities DataWorld-Bench reveals in existing agents, and whether training on separate synthesized worlds improves performance on held-out worlds and external benchmarks.

\subsection{Experimental Setup}

All construction stages use \textbf{GLM 5.3}~\citep{zhipu2026glm53} in the \textbf{Claude Code} harness~\citep{anthropic2026claudecode} (Section~\ref{sec:method}). The ten models in Table~\ref{tab:bench-results} use the same \textbf{mini-swe-agent} harness~\citep{swebench2025minisweagent} and the rubrics defined above, with a single judging pass over archived reports using Gemini-3.8-Flash as the rubric judge for both Tables~\ref{tab:bench-results} and~\ref{tab:base-vs-rl}. Unless otherwise noted, scores are reported on a $\times 100$ scale; DataWorld-Bench uses normalized rubric scores in $[0,1]$, while external benchmarks retain their native scoring. Higher is better. Model descriptions and harness details appear in Section~\ref{app:models}.

\subsection{DataWorld-Bench Results}

\begin{table*}[!t]
  \centering
  \footnotesize
  \setlength{\tabcolsep}{3.2pt}
  \noindent\begin{minipage}{\textwidth}
  \centering
  \captionof{table}{DataWorld-Bench results (normalized rubric scores $\times 100$, with SD in parentheses). Avg pools all rubric items and need not equal the mean of the two section columns. Best/second-best means are \textbf{bold}/\underline{underlined}. Rounds count model calls; tokens are prompt plus output of the final call (total) or outputs over the run (response).}
  \label{tab:bench-results}
  \resizebox{\textwidth}{!}{%
  \begin{tabular}{lcccccc}
    \toprule
    & \multicolumn{3}{c}{Score} & \multicolumn{3}{c}{Interaction and length} \\
    \cmidrule(lr){2-4} \cmidrule(l){5-7}
    Model & Data computation & Insight discovery & Avg & Rounds & Total tokens (k) & Resp.\ tokens (k) \\
    \midrule
    \modellogo{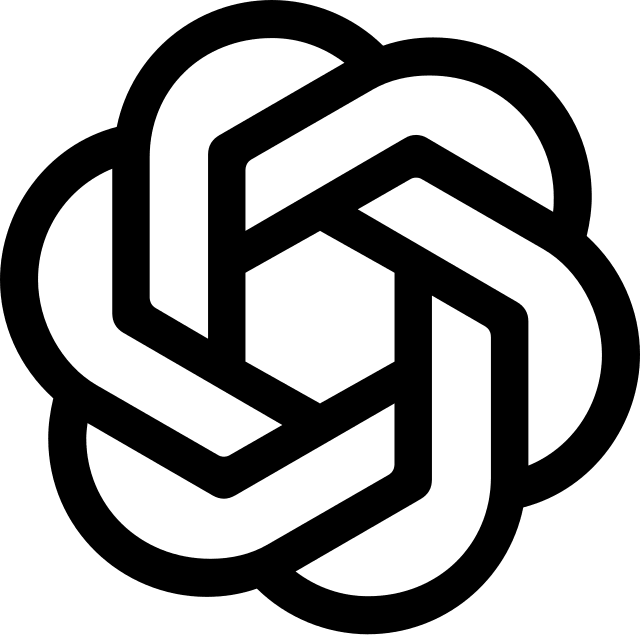}GPT-6-Astra         & $\mathbf{87.80}$\,\sd{16.38} & $\underline{66.45}$\,\sd{20.98} & $\mathbf{74.43}$\,\sd{15.85} & $28.7$\,\sd{8.7} & $121.9$k\,\sd{36.4} & $88.8$k\,\sd{18.9} \\
    \modellogo{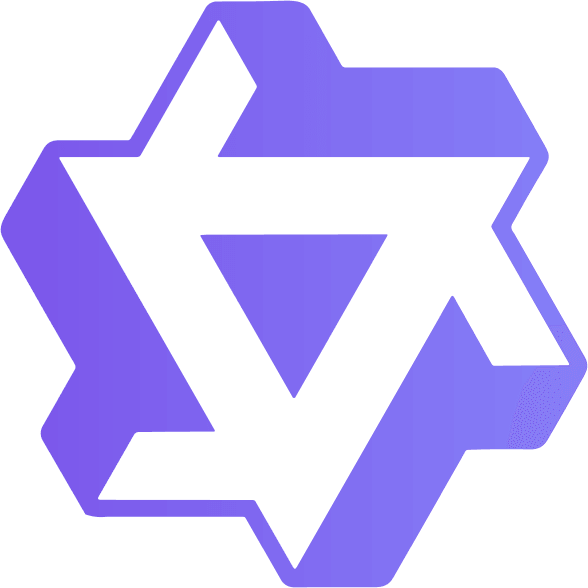}Qwen-3.8-Max-0902   & $82.36$\,\sd{17.79} & $\mathbf{67.95}$\,\sd{22.92} & $\underline{73.43}$\,\sd{16.81} & $82.7$\,\sd{23.5} & $289.7$k\,\sd{81.9} & $125.8$k\,\sd{33.0} \\
    \modellogo{openai}GPT-6.1-Sol         & $\underline{85.59}$\,\sd{17.26} & $66.02$\,\sd{21.02} & $73.30$\,\sd{16.44} & $31.2$\,\sd{10.6} & $135.2$k\,\sd{46.3} & $103.4$k\,\sd{30.4} \\
    \modellogo{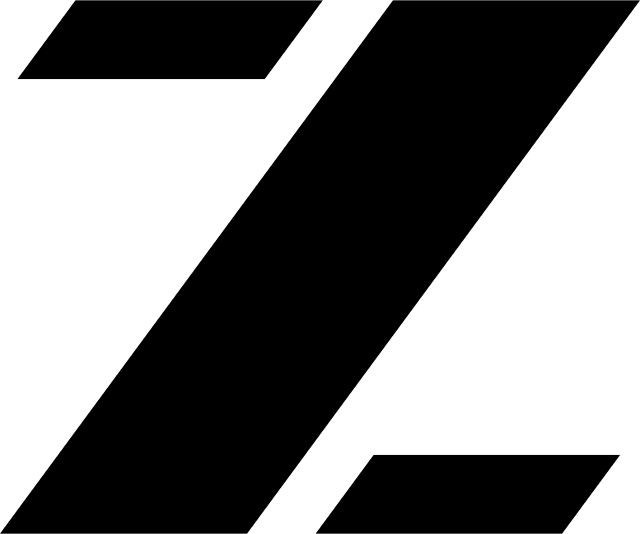}GLM 5.3             & $75.08$\,\sd{21.22} & $63.18$\,\sd{21.56} & $67.69$\,\sd{17.67} & $89.9$\,\sd{28.0} & $238.7$k\,\sd{96.6} & $116.3$k\,\sd{49.4} \\
    \modellogo{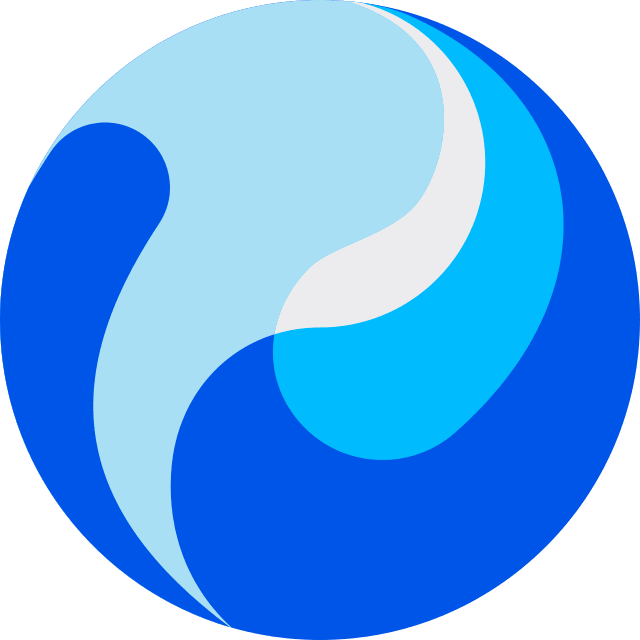}HY4-Preview         & $77.27$\,\sd{20.63} & $60.00$\,\sd{21.38} & $66.55$\,\sd{17.72} & $70.7$\,\sd{21.3} & $177.9$k\,\sd{49.2} & $86.0$k\,\sd{23.3} \\
    \modellogo{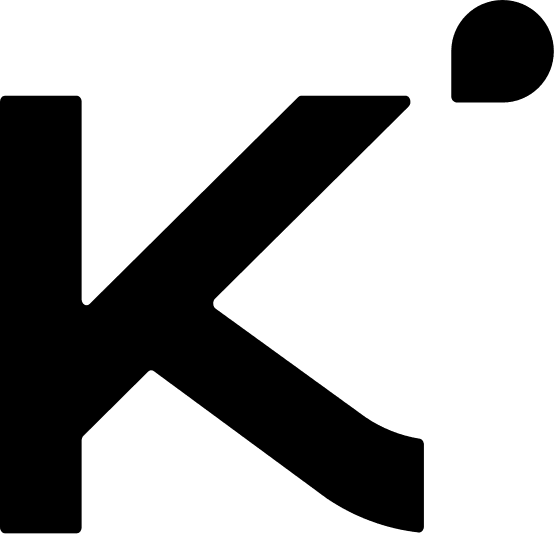}Kimi-K3             & $72.64$\,\sd{22.83} & $62.89$\,\sd{23.43} & $66.39$\,\sd{19.14} & $36.4$\,\sd{14.4} & $97.5$k\,\sd{35.1} & $48.0$k\,\sd{16.8} \\
    \modellogo{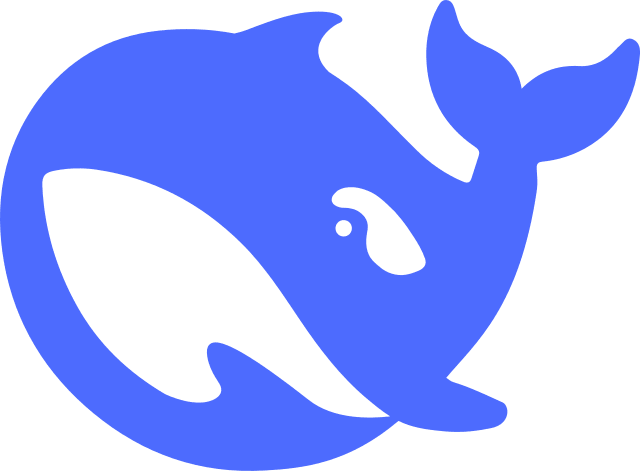}DeepSeek-V4.1-Flash & $72.30$\,\sd{19.72} & $57.34$\,\sd{20.29} & $62.95$\,\sd{16.93} & $58.1$\,\sd{20.2} & $150.7$k\,\sd{48.0} & $74.0$k\,\sd{24.3} \\
    \modellogo{qwen}Qwen-3.8-27B        & $66.93$\,\sd{25.33} & $54.93$\,\sd{22.46} & $59.64$\,\sd{20.16} & $55.1$\,\sd{18.5} & $170.4$k\,\sd{39.7} & $82.1$k\,\sd{20.7} \\
    \modellogo{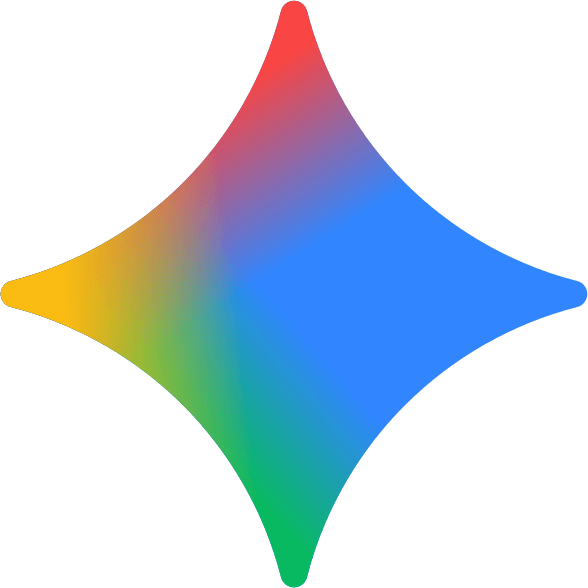}Gemini-3.8-Flash    & $59.00$\,\sd{21.70} & $54.13$\,\sd{21.18} & $56.17$\,\sd{17.01} & $62.3$\,\sd{15.6} & $98.8$k\,\sd{20.9} & $50.8$k\,\sd{10.3} \\
    \modellogo{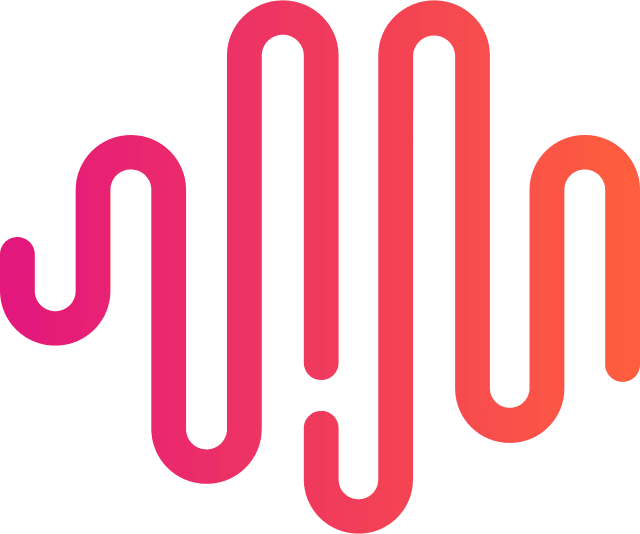}MiniMax-M3          & $53.39$\,\sd{20.44} & $47.39$\,\sd{19.85} & $49.75$\,\sd{16.89} & $93.8$\,\sd{32.1} & $158.4$k\,\sd{47.2} & $62.6$k\,\sd{19.4} \\
    \bottomrule
  \end{tabular}}%
  \end{minipage}
\end{table*}

\begin{table}[!t]
  \centering
  \small
  \setlength{\tabcolsep}{0pt}
  \caption{Distribution of triggered negative rubric items across trajectories.
    Each column gives the percentage of trajectories triggering exactly that many
    negative items; 3+ is omitted.}
  \label{tab:negative-rubric-items}
  \begin{tabular}{l@{\hspace{1.2em}}*{3}{>{\centering\arraybackslash}p{0.045\textwidth}}}
    \toprule
    & \multicolumn{3}{c}{Neg.\ rubric items (\%)} \\
    \cmidrule(lr){2-4}
    Model & 0 & 1 & 2 \\
    \midrule
    \modellogo{openai}GPT-6.1-Sol         & 86 & 11 & 2 \\
    \modellogo{openai}GPT-6-Astra         & 83 & 14 & 2 \\
    \modellogo{hunyuan}HY4-Preview         & 74 & 20 & 5 \\
    \modellogo{kimi}Kimi-K3             & 70 & 20 & 9 \\
    \modellogo{qwen}Qwen-3.8-Max-0902   & 68 & 26 & 5 \\
    \modellogo{zai}GLM 5.3             & 65 & 23 & 9 \\
    \modellogo{deepseek}DeepSeek-V4.1-Flash & 64 & 30 & 6 \\
    \modellogo{qwen}Qwen-3.8-27B        & 64 & 27 & 9 \\
    \modellogo{minimax}MiniMax-M3          & 59 & 28 & 10 \\
    \modellogo{gemini}Gemini-3.8-Flash    & 57 & 34 & 8 \\
    \bottomrule
  \end{tabular}
\end{table}

DataWorld-Bench remains challenging for frontier agents: the best overall and insight-discovery scores are 74.43 and 67.95, respectively (Table~\ref{tab:bench-results}). GPT-6-Astra leads overall and in computation, Qwen-3.8-Max-0902 leads in insight discovery and ranks second overall, and GPT-6.1-Sol ranks second in computation and third overall. Every model shows a computation--insight gap (4.87--21.35 points), indicating that strong data processing does not automatically yield targeted analytical conclusions. These comparisons concern rubric-defined skills, however, and criterion coverage and scoring differences may contribute to the gap.

Interaction cost does not uniformly track performance: GPT-6-Astra leads overall with the fewest rounds, while Kimi-K3 uses the fewest total and response tokens. Qwen-3.8-Max-0902 achieves the highest insight score with the largest token usage; GPT-6.1-Sol attains a similar overall score with fewer rounds and tokens. MiniMax-M3 takes the most rounds yet scores lowest, so longer interaction alone is not sufficient for stronger analysis. These cross-model comparisons do not isolate the benefit of additional inference, as models also differ in capability and interaction strategy.

Negative rubric items capture demonstrably wrong calculations or contradicted conclusions (Table~\ref{tab:negative-rubric-items}). GPT-6.1-Sol and GPT-6-Astra have the highest shares of trajectories without such items, while Gemini-3.8-Flash and MiniMax-M3 trigger them most often. Both GPT models trigger negative items less often than Qwen-3.8-Max-0902 but score lower in insight discovery, distinguishing avoidance of incorrect conclusions from recovery of targeted findings.

After removing each model's mean, domain analyses identify bioinformatics, operations, and energy as relatively easier and transportation, environment, and education as harder (Section~\ref{app:taxonomy}), motivating evaluation across diverse domains rather than aggregate scores alone; small per-domain task counts preclude within-domain model rankings. No model shows a detectable cause--consequence difficulty difference at the achieved resolution (Section~\ref{app:cause-cons}), providing no evidence that either graph route is systematically easier.

\subsection{Validating Training Utility}

\begin{table*}[t]
\centering
\footnotesize
\setlength{\tabcolsep}{3.2pt}
\noindent\begin{minipage}{\textwidth}
\centering
\captionof{table}{RL training utility: DataWorld-35B-A3B paired per task against its base Qwen3.6-35B-A3B (mean with SD in parentheses). Green/red badges indicate gains/losses relative to the base.}
\label{tab:base-vs-rl}
\resizebox{\textwidth}{!}{%
\begin{tabular}{lcccccccccccc}
\toprule
& \multicolumn{3}{c}{JobBench} & \multicolumn{3}{c}{DAComp} & \multicolumn{3}{c}{DABstep-Hard} & \multicolumn{3}{c}{DataWorld-Bench} \\
\cmidrule(lr){2-4} \cmidrule(lr){5-7} \cmidrule(lr){8-10} \cmidrule(l){11-13}
Model & Score & Rounds & Resp.\ tokens (k) & Score & Rounds & Resp.\ tokens (k) & Score & Rounds & Resp.\ tokens (k) & Score & Rounds & Resp.\ tokens (k) \\
\midrule
Qwen3.6-35B-A3B & $18.31$\,\sd{18.12} & $38.7$\,\sd{17.4} & $60.4$k\,\sd{22.1} & $+53.72$\,\sd{20.35} & $35.6$\,\sd{17.7} & $41.8$k\,\sd{19.6} & $+25.66$\,\sd{43.73} & $28.6$\,\sd{18.0} & $15.1$k\,\sd{15.3} & $37.64$\,\sd{11.62} & $36.5$\,\sd{13.9} & $38.4$k\,\sd{19.2} \\
DataMind-35B-A3B & \scorecell{22.93}{22.33}{4.63} & $36.6$\,\sd{12.7} & $73.9$k\,\sd{28.0} & \scorecell{+54.79}{20.49}{1.07} & $37.6$\,\sd{14.1} & $47.7$k\,\sd{19.0} & \scorecell{+31.22}{46.40}{5.56} & $40.3$\,\sd{17.6} & $26.3$k\,\sd{16.8} & \scorecell{37.36}{11.14}{-0.29} & $34.4$\,\sd{9.6} & $42.0$k\,\sd{17.2} \\
BigBang-v1 & \scorecell{20.84}{20.69}{2.53} & $68.9$\,\sd{23.4} & $84.3$k\,\sd{30.1} & \scorecell{+54.31}{20.93}{0.58} & $45.1$\,\sd{22.9} & $38.2$k\,\sd{18.4} & \scorecell{+34.66}{47.65}{8.99} & $37.2$\,\sd{24.5} & $35.5$k\,\sd{34.0} & \scorecell{40.05}{12.06}{2.41} & $58.4$\,\sd{27.8} & $46.7$k\,\sd{18.1} \\
Occamy-1.0 & \scorecell{30.18}{23.79}{11.87} & $79.2$\,\sd{41.7} & $96.4$k\,\sd{38.9} & \scorecell{+64.86}{20.89}{11.13} & $49.1$\,\sd{22.5} & $45.4$k\,\sd{21.6} & \scorecell{+44.44}{49.76}{18.78} & $31.0$\,\sd{21.3} & $26.3$k\,\sd{28.4} & \scorecell{42.38}{14.06}{4.74} & $61.5$\,\sd{25.8} & $58.2$k\,\sd{28.3} \\
\rowcolor{rowblue}
DataWorld-35B-A3B & \scorecell{30.43}{23.92}{12.13} & $63.6$\,\sd{28.9} & $88.1$k\,\sd{29.1} & \scorecell{+68.00}{18.90}{14.28} & $78.4$\,\sd{25.0} & $65.9$k\,\sd{21.7} & \scorecell{+46.56}{49.95}{20.90} & $43.2$\,\sd{22.3} & $36.3$k\,\sd{28.0} & \scorecell{44.70}{13.94}{7.06} & $102.8$\,\sd{22.6} & $68.4$k\,\sd{17.6} \\
\bottomrule
\end{tabular}}%
\end{minipage}
\end{table*}

\begin{table}[t]
  \centering
  \small
  \setlength{\tabcolsep}{5pt}
  \caption{General-capability and factuality checks for DataWorld-35B-A3B versus Qwen3.6-35B-A3B. MMLU-Pro and LiveCodeBench (LCB)-Hard report accuracy (\%); SimpleQA-Verified (SimpleQA-V) reports F1. Higher is better.}
  \label{tab:general-capabilities}
  \resizebox{\columnwidth}{!}{%
  \begin{tabular}{@{}lccc@{}}
    \toprule
    Model & MMLU-Pro & LCB-Hard & SimpleQA-V \\
    \midrule
Qwen3.6-35B-A3B & 84.73 & 71.14 & 22.20 \\
\rowcolor{rowblue}
DataWorld-35B-A3B & 84.73 & 70.29 & 24.00 \\
    \bottomrule
  \end{tabular}}%
\end{table}

We assess the second question by using the engine's separate training set for RL and measuring gains on world-disjoint DataWorld-Bench tasks and three external benchmarks. The held-out benchmark measures training outcomes; its worlds and answers are not training inputs.
We train Qwen3.6-35B-A3B~\citep{qwen36_35b_a3b} on the training split in Section~\ref{sec:curation} using Group Sequence Policy Optimization (GSPO; \citealp{zheng2025gspo}) in an RL-zero setting. Construction models supply environments, tasks, and rubrics but no task-solving demonstrations; optimization uses only the training model's own rollouts and rubric rewards, avoiding task-specific supervised fine-tuning or stronger-model trajectory distillation as confounds. To reduce RL reward-scoring costs, we use the Qwen3.6-35B-A3B base model itself as the rubric judge. For a task $x$, let $y_i$ be one of $G$ sampled responses with rubric reward $r_i$. GSPO computes a group-normalized advantage $\widehat A_i=(r_i-\bar r)/\operatorname{std}(r_1,\ldots,r_G)$ and a length-normalized, sequence-level importance ratio $s_i(\theta)=\exp\big(\tfrac{1}{|y_i|}\sum_{t=1}^{|y_i|}\log\rho_{i,t}(\theta)\big)$, where $\bar r$ is the group mean reward and $\rho_{i,t}(\theta)$ is the ratio of $\pi_\theta$ to the sampling policy $\pi_{\rm old}$ at token $t$, conditioned on $x$ and $y_{i,<t}$. Its clipped objective is
\begin{equation}
\begin{aligned}
\mathcal{J}(\theta)&=\operatorname{E}_{x,\{y_i\}\sim\pi_{\rm old}}
\left[\frac{1}{G}\sum_{i=1}^{G}\min\!\left(
 s_i(\theta)\widehat A_i,\right.\right.\\
&\hspace{1em}\left.\left.\operatorname{clip}(s_i(\theta),1-\epsilon_{\rm low},1+\epsilon_{\rm high})
\widehat A_i\right)\right],
\end{aligned}
\end{equation}
\begin{figure}[t]
  \centering
  \includegraphics[width=\columnwidth]{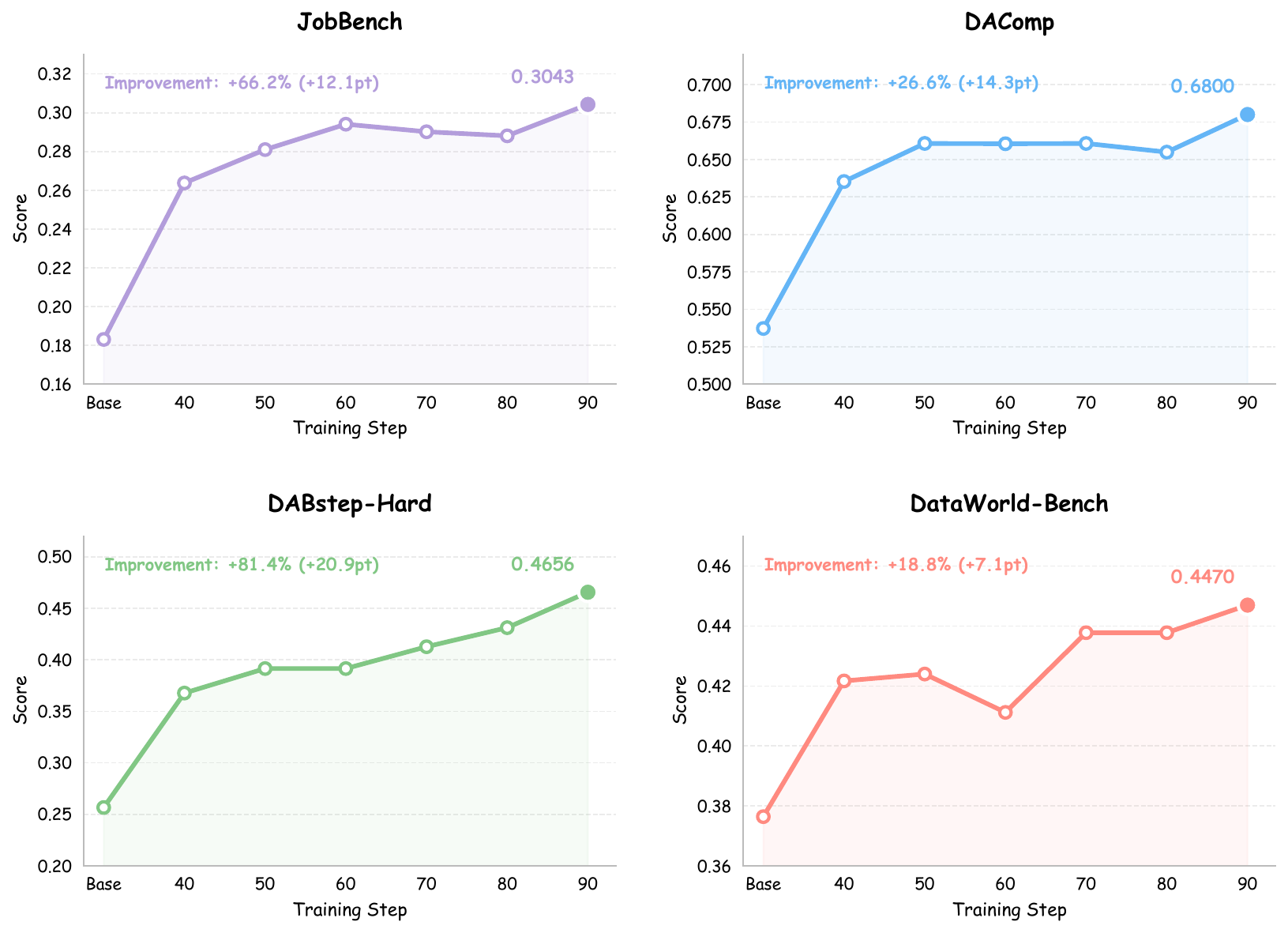}
  \caption{Performance as RL training scales over synthesized tasks. Base is the pre-RL model; scores are unscaled, with final gains over Base annotated in points and, where applicable, relative percentages.}
  \label{fig:benchmark-improvement}
\end{figure}
where $\epsilon_{\rm low}$ and $\epsilon_{\rm high}$ are the lower and upper clipping thresholds; the same ratio and advantage apply to the entire response. Training implementation details appear in Section~\ref{app:rl-training}.

\paragraph{Comparisons.}
Table~\ref{tab:base-vs-rl} compares \textbf{DataWorld-35B-A3B} with its base and three models sharing that base. All use the same mini-swe-agent evaluation protocol, with identical tasks and benchmark-specific scoring (Section~\ref{app:rl-eval}). For a fair comparison, we retrain \textbf{DataMind-35B-A3B} on 7{,}323 publicly released Python data-analysis tasks~\citep{qiao2025scaling} using our backbone and training configuration, since the original model uses a different backbone and a context window too short for our evaluation. We also evaluate the released BigBang-v1~\citep{endlessfrontier2026bigbang} and Occamy-1.0~\citep{acciolab2026occamy} checkpoints, which share our backbone and have undergone extensive multi-stage training. These baselines use teacher-distilled trajectories. The comparisons assess overall training utility rather than isolate the effects of training data or recipes.

\paragraph{Results.}
DataWorld-35B-A3B achieves the highest scores among the evaluated same-base models on all four benchmarks (Table~\ref{tab:base-vs-rl}). It gains 7.06 points over the base on DataWorld-Bench, supporting transfer to world-disjoint tasks rather than reuse of training-world answers. Gains of 14.28 points on DAComp~\citep{lei2026dacomp} and 20.90 on DABstep-Hard~\citep{egg2025dabstep} extend to open-ended analysis and automatically checked answers, respectively. Notably, JobBench~\citep{li2026jobbench} covers 35 occupations, not just data analysis, making it the least task-aligned of the four suites. The 12.13-point gain on its main set supports transfer beyond data analysis to broader file-based professional reasoning. In contrast, DataMind improves the external benchmarks but leaves held-out DataWorld-Bench essentially unchanged, suggesting that gains on existing benchmarks need not extend to mechanism-grounded insight tasks. The joint increases in scores, rounds, and response tokens across all four benchmarks suggest stronger \emph{long-horizon reasoning and tool interaction}. This long-context, multi-turn interaction pattern resembles that documented in MiMo's public RL training logs~\citep{xiaomi2026mimorl}.

Figure~\ref{fig:benchmark-improvement} connects these gains to the supply bottleneck that motivates Data World: as RL training draws successive batches from the synthesized task set, scores on all four benchmarks rise from the pre-RL base to their final levels. These trajectories highlight the training value of combining synthesized tasks with mechanism-grounded rewards: the resulting improvements extend beyond the generated environments to external analytical and professional workflows.

\paragraph{General-capability retention and factuality.}
Table~\ref{tab:general-capabilities} probes general knowledge (MMLU-Pro; \citealp{wang2024mmlupro}), algorithmic coding (LiveCodeBench-Hard; \citealp{jain2024livecodebench}), and factual question answering (SimpleQA-Verified; \citealp{haas2025simpleqaverified}). MMLU-Pro accuracy is unchanged at 84.73\%; LiveCodeBench-Hard accuracy decreases from 71.14\% to 70.29\%, while SimpleQA-Verified F1 increases from 22.20 to 24.00. These results indicate that RL training on fully synthesized environments preserves the base model's general language and coding capabilities, and that purely synthetic supervision does not noticeably disturb its factual world knowledge.

\section{Conclusion}

Data World offers a data-centric route to scalable, verifiable supervision for data agents by specifying generative mechanisms before synthesizing observations and auditing analytical targets. Separating canonical truth from analyst-visible files grounds supervision in known mechanisms while requiring agents to recover findings from observations alone. The engine yields DataWorld-Bench for measuring the computation--insight gap and a world-disjoint set for RL-zero training; gains extend to unseen worlds, external data-analysis tasks, and broader occupational workflows, without task-solving demonstrations from stronger models. This joint synthesis expands task supply without requiring curators to rediscover each targeted finding from the generated files. Together, these results support mechanism-grounded environment synthesis as a scalable complement to real-world data, providing both evaluation criteria and training signals.

\ifshowconferenceendmatter
\section*{Limitations}

First, the tight coupling among synthesis components and the cost of regenerating worlds and tasks limit exhaustive ablations. We do not construct multiple controlled variants of the world and task synthesis pipeline, so our results support the utility of the framework as a whole rather than isolate the contribution of each component. Moreover, the central claim concerns what synthesis from scratch enables---verifiable, scalable supervision---rather than the optimality of any particular component, so adding, removing, or fine-tuning individual components is unlikely to yield a materially better outcome: weakening the pipeline's coherence would sacrifice the auditability and coverage that motivate the design. Synthesized environments and tasks are intended to complement, not replace, real-world data; we do not compare which source is more effective.

Second, fully synthetic tasks may differ from the questions human users pose to AI agents in practice. Although human preference-based curation selects benchmark questions for realism, it does not establish that the resulting tasks represent the distribution of real user requests, particularly their ambiguity, implicit context, and evolving goals.

Third, mechanism-grounded rubrics do not exhaust all valid analytical findings, and LLM-based grading may introduce judge bias. Audited generative truth makes the specified criteria verifiable, but does not guarantee that alternative insights receive appropriate credit or that scores remain consistent across different judging models.

\section*{Ethical Considerations}

Environment synthesis and scaling for agent training have recently become active research areas, with many concurrent contributions appearing in rapid succession. Due to page constraints and the limits of our knowledge, we cite and discuss only the most closely related work known to us; we regret any relevant omissions. AI agents construct environments and rubrics, and an LLM judge scores responses under the access restrictions described in Section~\ref{sec:method}. These artifacts and scores do not establish suitability for real-world decisions: audits verify specified relationships and recoverability, while human curation assesses question realism rather than operational validity. Although synthesis reduces reliance on private records, outputs should still be reviewed for sensitive or proprietary content before release. Use in consequential domains requires independent validation and human oversight. AI tools were also used to assist with code development and manuscript writing; the authors reviewed, edited, and validated all generated content and take full responsibility for the correctness of the code and the manuscript.
\fi

\bibliography{custom}

\appendix

\section{Appendix}
\label{app:overview}

This appendix provides an intuitive theoretical analysis of rubric coverage (Section~\ref{app:intuitive-theory}), introduces the evaluated models (Section~\ref{app:models}), describes the training and test distributions and additional benchmark analyses (Section~\ref{app:benchmark-analysis}), and documents the RL training configuration (Section~\ref{app:rl-training}) and evaluation protocol (Section~\ref{app:rl-eval}).

\subsection{Intuitive Theoretical Analysis}
\label{app:intuitive-theory}

For a fixed task and observed environment $\mathcal{E}$, let $S$ be the set of valid atomic findings recoverable from $\mathcal{E}$, and let $R_{\mathrm{obs}}$ and $R_{\mathrm{DW}}$ denote the findings covered by positive rubric items constructed from observed files alone and with access to the Data World generative record, respectively. Findings are compared by their meaning rather than wording; negative penalty items are excluded. The overlap $|S\cap R|$ measures the number of valid findings with matching evaluation criteria.

Crucially, our question-writing agent can see the original world specification, including the mechanisms and target phenomena established before data generation, as well as the generated state and audit artifacts. It need not rediscover these specified phenomena solely by exploring files. This information provides candidate findings and verification evidence that can reduce discovery-related omissions.

Under a matched rubric-item budget, assume an idealized writer can reproduce the file-only strategy, ignore unhelpful additional information, and choose the rubric with the greatest valid-finding coverage among its available strategies. Writing $R_{\mathrm{obs}}^{\star}$ and $R_{\mathrm{DW}}^{\star}$ for these idealized choices, the inclusion of the file-only strategy gives
\begin{equation}
|S\cap R_{\mathrm{DW}}^{\star}|
\geq |S\cap R_{\mathrm{obs}}^{\star}|.
\end{equation}
A strict improvement is possible when specification access enables additional valid findings to be covered within the same budget. Candidates must still be checked for recoverability from $\mathcal{E}$: a fact known from the hidden state alone need not belong to $S$. This inequality describes an idealized information advantage, not a guaranteed ordering of actual LLM-generated rubrics or exhaustive coverage of valid findings.

\subsection{Evaluated Models}
\label{app:models}

We evaluate ten models on DataWorld-Bench (Table~\ref{tab:bench-results}). The descriptions below summarize their public technical reports, model cards, and system cards. All ten use the mini-swe-agent harness~\citep{swebench2025minisweagent}, a lightweight harness that exposes bash as its sole tool and follows a ReAct-style loop, interleaving model reasoning, shell-command execution, and feedback from command outputs.

\paragraph{GPT-6-Astra}~\citep{openai2026gpt6astra} is OpenAI's flagship reasoning model in the GPT-6 family for complex, tool-using workflows. Its system card describes reinforcement-learning-based reasoning training and improved robustness to prompt injection and unauthorized actions in agentic settings.

\paragraph{GPT-6.1-Sol}~\citep{openai2026gpt61sol} is OpenAI's efficiency-oriented model in the GPT-6.1 family. OpenAI reports capabilities comparable to GPT-6 Astra with improved speed and affordability; its system-card addendum states that it uses the same types of data and training as Astra.

\paragraph{Qwen-3.8-Max-0902}~\citep{alibaba2026qwen38max0902} is Alibaba's largest Qwen model: a 2.4T-parameter mixture-of-experts with 95B activated per token, a 1M-token context window, and multimodal input (text, image, and video). The 0902 snapshot (released 2026-09-02) upgrades the August Qwen3.8-Max release specifically for coding and agent performance; Alibaba reports fifth place in Text Arena and second in Vision Arena for the base release, and its open weights have become a popular base for coding agents.

\paragraph{Kimi-K3}~\citep{moonshot2026kimik3} is Moonshot AI's open-weight flagship and the first open 3T-class model: a 2.8T-parameter mixture-of-experts with 896 experts (top-16 routing), KDA attention, a 1M-token context window, and native multimodality. Its technical report documents frontier-level performance across long-horizon coding, agentic, knowledge, and reasoning evaluations; third-party tracking places it at \#3 on the Artificial Analysis Intelligence Index~\citep{artificialanalysis2026index}, and it attains an Elo of 1668 on GDPval-AA v2~\citep{patwardhan2025gdpval} for agentic office work.

\paragraph{HY4-Preview}~\citep{tencent2026hy4preview} is Tencent Hunyuan's most capable open model to date: a 770B-parameter MoE with 49B activated parameters (256 routed experts plus one shared expert, top-8 routing) and a 1M-token context window, released under Apache 2.0. Tencent reports its largest generation-over-generation gain so far, with the biggest improvements on long-horizon software engineering, document-heavy office work, and scientific research. In blind evaluations of 203 real engineering tasks rated by 163 experts, HY4-Preview averaged 2.99/4.00, edging out GLM 5.3 (2.92; 46.8\% wins vs.\ 40.4\% losses) and Kimi-K3 (2.94; 51.2\% wins vs.\ 40.9\% losses).

\paragraph{GLM 5.3}~\citep{zhipu2026glm53} is Zhipu AI's (Z.ai) coding-focused flagship: a 743B model sharing the GLM-5.2 backbone with a substantially upgraded post-training run. Z.ai reports it as the most capable open-weights model for coding, with a 50\% improvement over GLM-5.2 on its in-house Z.ai Code Bench, and highlights emergent defensive-cybersecurity capabilities alongside strong performance on long-horizon autonomous engineering tasks.

\paragraph{DeepSeek-V4.1-Flash}~\citep{deepseek2026v41flash} targets extreme long-context serving efficiency: a 552B backbone built on a Causal Encoder-Decoder (CED) architecture with 8B prefill and 16B decode activated parameters and a 1M-token context. Its technical report pushes KV cache compression across architecture, numeric precision, and deployment jointly, reducing the global (always-in-HBM) KV cache footprint to 890 bytes per token---about $4\times$ smaller than DeepSeek-V4-Flash and $437\times$ smaller than DeepSeek-V1.

\paragraph{Gemini-3.8-Flash}~\citep{google2026gemini38flash} is Google DeepMind's lightweight, API-only entry in the Gemini 3.8 family. Despite its size class, it is reported to lead Harvey's Legal Agent Benchmark and to score 90.8\% on Terminal-Bench 2.1~\citep{laude2025terminalbench} (vs.\ 81.6\% for its 3.7 Flash predecessor); a companion Flash Cyber variant improves recall by 7.5--9.7\% on an internal penetration-testing benchmark at 2.3--5.2$\times$ lower cost. No formal technical report has been published.

\paragraph{Qwen-3.8-27B}~\citep{alibaba2026qwen3827b} is Alibaba Qwen's dense 27B natively multimodal model (64 layers, hidden size 5120), combining a Gated DeltaNet and Gated Attention hybrid architecture with a 262K native context window extensible to 1M tokens via YaRN~\citep{peng2024yarn}, released under Apache 2.0. Third-party analyses report it competitive with models $10$--$15\times$ its size on coding, reasoning, multimodal, and agentic benchmarks while remaining deployable on a single 24GB GPU.

\paragraph{MiniMax-M3}~\citep{lai2026minimax} is MiniMax's production-grade natively multimodal model built on MiniMax Sparse Attention (MSA). MSA performs on par with GQA while reducing per-token attention compute by $28.4\times$ at 1M context, delivering $9\times$ prefill and $15\times$ decode efficiency gains; MiniMax reports frontier-level performance on specialized coding and agentic workloads.

\subsection{Benchmark Composition and Additional Analyses}
\label{app:benchmark-analysis}

\subsubsection{Training and Test Distributions}
\label{app:split-distribution}

Figure~\ref{fig:split-distribution} compares task shares across the 16 taxonomy categories in the held-out DataWorld-Bench test split and the Data World RL training set. Percentages are normalized within each split. The test set covers the categories approximately uniformly, whereas the training set has a more uneven distribution; the figure describes coverage, not task difficulty.

\subsubsection{Task Case Studies}
\label{app:task-cases}

We present three English task prompts together with their complete analyst-visible file inventories to illustrate task diversity and cross-system analytical requirements. Each environment directory (\texttt{env/}) is mounted at \texttt{/app/data/} in the sandbox; directory cards preserve the supplied file names and size labels. We omit runner-injected workflow instructions and replace the request prefix with a display label. These examples are task inputs, not model solutions or evidence of successful task completion.

\paragraph{Public safety and deployment (cause route).}
This environment represents a midsize U.S. city with 47 census tracts over 2015--2025, combining police, school, youth-service, housing-court, public-health, and juvenile-justice records. The task requires linking encounters to deployment and budgets, reconciling geographic and identifier crosswalks, and assessing volume and demographic patterns while accounting for the distribution of policing.
\begin{taskenv}
\envdir{census}
\envfile{acs_pums_households.csv}{11.7M}
\envfile{acs_tract_estimates.csv}{44K}
\envfile{acs_variable_codebook.pdf}{3K}
\envfile{housing_court_filings.csv}{5.1M}
\envfile{snap_enrollment_records.csv}{10.5M}
\envfile{tract_boundary_crosswalk_2020.csv}{394B}
\envdir{crosswalks}
\envfile{tract_precinct_mapping.csv}{3K}
\envfile{youth_student_id_crosswalk.csv}{46K}
\envdir{health}
\envfile{hospital_network_directory.csv}{392B}
\envfile{icd_coding_reference.pdf}{2K}
\envfile{youth_injury_er.csv}{432K}
\envdir{juvenile_justice}
\envfile{case_records.csv}{922K}
\envfile{disposition_code_reference.csv}{296B}
\envfile{youth_id_crosswalk_2019.csv}{20K}
\envdir{municipal}
\envfile{budget_allocations.csv}{102K}
\envfile{budget_vs_actual_variance.csv}{38K}
\envfile{program_code_crosswalk.csv}{5K}
\envfile{youth_program_quarterly_reports.csv}{80K}
\envfile{youth_program_site_directory.csv}{5K}
\envdir{police}
\envfile{cad_rms_codebook.pdf}{2K}
\envfile{deployment_monthly.csv}{73K}
\envfile{encounters_cad.csv}{21.3M}
\envfile{precinct_boundary_crosswalk_2019.csv}{344B}
\envfile{violence_incidents.csv}{2.4M}
\envdir{school}
\envfile{attendance_methodology_memo_sy2018.pdf}{3K}
\envfile{school_directory.csv}{1K}
\envfile{student_engagement_semester.csv}{25.8M}
\envdir{surveys}
\envfile{community_trust_biennial.csv}{29K}
\envfile{survey_methodology_documentation.pdf}{2K}
\end{taskenv}
\begin{taskprompt}
I sit on the Public Safety Committee, and every month we get the CAD encounter dump and a one-line ``stops are up X\%''---and that's the whole analysis. I want to know what's really behind the volume and where it's falling. Is it how many officers we have on the street, the budget, the neighborhoods we police, the time of year? When patrol goes up, do stops go up proportionally, or do they plateau? And the thing I keep getting asked by constituents: are stops and searches falling harder on Black neighborhoods and Black residents, and is that about race or about where we deploy?

Pull the encounter records together with the deployment and budget data and give me a clear picture of what predicts encounter volume, where the encounters concentrate and why, and whether the racial pattern holds up once you account for how policing is distributed. I need to take this into a budget hearing, so tell me what's a real driver, what's a data artifact, and what looks like a pattern but dissolves on closer look.
\end{taskprompt}

\paragraph{Class-capacity planning (cause route).}
This environment models a Los Angeles prenatal-yoga and maternal-community platform with three locations, combining scheduling, billing, customer, community, pricing, and market records. The task centers on annual scheduling exports and market intelligence to compare seasonal and cross-site capacity patterns and assess competing explanations for them.
\begin{taskenv}
\envdir{billing}
\envfile{ltv_monthly_snapshot.csv}{2.6M}
\envfile{orders_transactions.csv}{874K}
\envfile{subscription_events.csv}{566K}
\envdir{community}
\envfile{community_guidelines.txt}{1001B}
\envfile{daily_user_activity.parquet}{284K}
\envfile{mom_group_directory.csv}{2K}
\envfile{posts_raw_sample.jsonl}{561K}
\envdir{crm}
\envfile{health_score_assessments.csv}{379K}
\envfile{nps_survey_responses.csv}{127K}
\envfile{user_current_view.csv}{124K}
\envfile{user_profiles_scd.csv}{801K}
\envdir{market}
\envfile{competitor_scan_notes.txt}{1K}
\envfile{monthly_market_index.csv}{1K}
\envdir{revenue}
\envfile{discount_rules_history.csv}{2K}
\envfile{price_adjustment_log.csv}{21K}
\envfile{pricing_tiers.xlsx}{11K}
\envfile{service_menu_2023.pdf}{2K}
\envdir{scheduling}
\envfile{class_sessions_2022.json}{270K}
\envfile{class_sessions_2023.json}{343K}
\envfile{class_sessions_2024.json}{384K}
\envfile{instructor_roster.csv}{1K}
\envfile{room_capacity_ref.csv}{266B}
\end{taskenv}
\begin{taskprompt}
Hi---Maya from operations here. We scale class capacity up and down every month and honestly nobody on the team can agree on what's actually behind it. Marketing is convinced we react to competitor promotions; the scheduling team swears we ramp up ahead of the summer baby boom. On top of that, our three locations never seem to carry the same capacity and I'd like to understand why.

Can you dig into the scheduling exports and the market intel file and tell me what's really driving our scheduled class capacity---both the month-to-month swings and why the sites differ---and whether there's anything that lets us see the swings coming? I need something we can actually plan against, not just a description of what happened.
\end{taskprompt}

\paragraph{Production yield and downstream availability (consequence route).}
This environment follows the Hillcats' 2026 game-day supply operations, linking plant utilization and quality inspections with procurement, warehouse inventories, shipments, and ballpark availability. The task traces whether production-related yield losses coincide with downstream shortages or are absorbed by warehouse buffers, using operational records and contextual memos.
\begin{taskenv}
\envfile{00_DATA_ROOM_README.txt}{7K}
\envdir{finance}
\envfile{game_day_sales_summary_2026.csv}{5K}
\envdir{gameday}
\envfile{supply_availability_2026.csv}{222K}
\envdir{memos}
\envfile{memo_2026-06-16_midseason_count_riskscore_v2.txt}{1K}
\envfile{memo_2026-08-01_supplier_change.txt}{1K}
\envdir{production}
\envfile{line_utilization_daily_2026.csv}{7K}
\envfile{shift_supervisor_notes_2026.txt}{9K}
\envdir{qms}
\envfile{batch_inspections_2026.csv}{23K}
\envdir{reference}
\envfile{defect_codebook.csv}{1K}
\envfile{facility_codes.csv}{1K}
\envfile{sku_catalog.csv}{4K}
\envdir{schedule}
\envfile{attendance_forecast_revisions_2026.csv}{27K}
\envfile{hillcats_schedule_2026.csv}{12K}
\envfile{promo_calendar_2026.json}{4K}
\envdir{scm}
\envfile{purchase_order_lines_2026.csv}{133K}
\envfile{supplier_master_scd2_2026.xlsx}{7K}
\envdir{weather}
\envfile{station_metadata_and_notes.txt}{2K}
\envfile{weather_daily_klyh_2026.csv}{8K}
\envdir{wms}
\envfile{inventory_movements_2026.csv.gz}{107K}
\envfile{inventory_snapshot_sku_wh_2026.csv.gz}{145K}
\envfile{physical_count_2026-06-17.csv}{4K}
\envfile{po_receipts_2026.csv}{120K}
\envfile{shipments_2026.csv}{333K}
\end{taskenv}
\begin{taskprompt}
I run Gameday Supply ops for the Hillcats, and I owe the front office a season postmortem on our central plant. Twice this year we ran the shop flat-out for days on end---the stretch leading into Opening Day, and again a shorter one in mid-June---and the shift supervisors kept telling me quality goes to hell when we sprint like that. I need to know whether that's actually in the numbers: does pushing the lines hard cost us yield, how much, and under exactly what conditions? And the part the front office will actually care about: did those quality losses reach the ballpark---did fans end up staring at empty shelves because of how we ran the plant---or did it all get absorbed in the warehouses somewhere? Quantify whatever you can, be straight with me about what the data can and can't support, and give me something I can defend in the season review.
\end{taskprompt}

\subsubsection{Per-Domain Results}
\label{app:taxonomy}

\begin{table*}[t]
  \centering
  \scriptsize
  \setlength{\tabcolsep}{2.6pt}
  \caption{Per-domain scores, following Table~\ref{tab:bench-results}. Avg pools all 100 test tasks. The bottom row averages domain offsets after removing each model's mean. Each domain has only $6$--$8$ tasks, so individual cells should not be used to rank models. Domain abbreviations are defined in Section~\ref{app:taxonomy}.}
  \label{tab:bench-taxonomy}
  \resizebox{\textwidth}{!}{%
  \begin{tabular}{lccccccccccccccccc}
    \toprule
    & \multicolumn{16}{c}{Taxonomy category} & \\
    \cmidrule(lr){2-17}
    Model & Com & Cus & Fin & Ops & HR & Risk & Clin & Epi & Bio & Sci & Env & Enrg & Pol & Edu & Soc & Tran & Avg \\
    \midrule
    \modellogo{openai}GPT-6-Astra          & 69.96 & 77.25 & 79.79 & 84.91 & 65.68 & 73.12 & 71.73 & 70.54 & 83.31 & 85.53 & 64.22 & 77.27 & 68.46 & 72.65 & 77.60 & 69.77 & \textbf{74.43} \\
    \modellogo{qwen}Qwen-3.8-Max-0902    & 78.58 & 81.23 & 73.15 & 80.35 & 77.81 & 65.23 & 76.70 & 73.68 & 76.08 & 77.98 & 60.82 & 88.12 & 68.99 & 62.23 & 76.15 & 58.28 & 73.43 \\
    \modellogo{openai}GPT-6.1-Sol          & 76.20 & 77.60 & 79.59 & 76.93 & 57.45 & 69.65 & 74.14 & 69.81 & 74.41 & 92.71 & 66.50 & 80.69 & 68.25 & 67.10 & 75.00 & 67.90 & 73.30 \\
    \modellogo{zai}GLM 5.3              & 71.95 & 69.50 & 71.62 & 75.45 & 73.97 & 69.53 & 62.54 & 68.52 & 78.43 & 68.81 & 50.27 & 75.91 & 64.35 & 58.02 & 75.16 & 47.68 & 67.69 \\
    \modellogo{hunyuan}HY4-Preview          & 67.65 & 64.90 & 76.25 & 73.26 & 65.97 & 54.39 & 63.98 & 62.10 & 70.51 & 68.05 & 65.86 & 79.36 & 62.50 & 59.23 & 75.29 & 53.88 & 66.55 \\
    \modellogo{kimi}Kimi-K3              & 61.79 & 69.68 & 74.68 & 78.24 & 64.06 & 55.31 & 65.44 & 75.03 & 70.48 & 70.09 & 63.32 & 72.96 & 59.84 & 63.58 & 68.88 & 50.23 & 66.39 \\
    \modellogo{deepseek}DeepSeek-V4.1-Flash  & 69.93 & 66.05 & 60.35 & 69.83 & 64.84 & 61.10 & 60.15 & 56.79 & 74.99 & 65.96 & 50.74 & 66.64 & 61.07 & 57.04 & 66.23 & 55.04 & 62.95 \\
    \modellogo{qwen}Qwen-3.8-27B         & 53.43 & 70.51 & 62.47 & 70.76 & 67.29 & 61.11 & 51.31 & 65.13 & 69.58 & 52.27 & 49.73 & 53.58 & 60.09 & 55.13 & 60.71 & 50.55 & 59.64 \\
    \modellogo{gemini}Gemini-3.8-Flash     & 58.83 & 55.95 & 52.75 & 60.86 & 55.82 & 43.48 & 62.52 & 57.09 & 72.78 & 47.79 & 58.40 & 59.46 & 52.09 & 43.64 & 70.29 & 43.65 & 56.17 \\
    \modellogo{minimax}MiniMax-M3           & 54.94 & 53.08 & 44.04 & 52.14 & 44.95 & 47.40 & 59.15 & 53.65 & 56.67 & 51.48 & 35.43 & 58.37 & 47.88 & 49.44 & 49.23 & 38.90 & 49.75 \\
    \midrule
    $n$ (tasks) & 6 & 6 & 6 & 6 & 6 & 6 & 6 & 6 & 6 & 6 & 6 & 6 & 8 & 6 & 8 & 6 & 100 \\
    Dev. from model mean & $+1.30$ & $+3.55$ & $+2.44$ & $+7.25$ & $-1.25$ & $-5.00$ & $-0.27$ & $+0.21$ & $+7.70$ & $+3.04$ & $-8.50$ & $+6.21$ & $-3.68$ & $-6.22$ & $+4.43$ & $-11.44$ & $0.00$ \\
    \bottomrule
  \end{tabular}}%
\end{table*}

\paragraph{Domain key.}
Com: Commercial Growth \& Acquisition; Cus: Customer, Pricing \& Revenue; Fin: Finance, FP\&A \& Markets; Ops: Operations, Supply Chain \& Manufacturing; HR: People; Risk: Risk, Fraud \& Compliance; Clin: Clinical \& Healthcare; Epi: Epidemiology \& Public Health; Bio: Bioinformatics \& Omics; Sci: Experimental \& Physical Science; Env: Environment, Climate \& Geospatial; Enrg: Energy, Utilities \& IoT; Pol: Public Policy \& Program Evaluation; Edu: Education \& Learning Analytics; Soc: Social Science, Survey \& Behavioral; Tran: Transportation \& Mobility. These categories correspond to each source world's \texttt{dag\_category} label.

\paragraph{Results.}
Table~\ref{tab:bench-taxonomy} breaks scores down by the 16 taxonomy categories, with Avg matching Table~\ref{tab:bench-results}. The bottom row removes each model's mean before averaging across all ten models to describe relative domain difficulty. Bioinformatics \& omics ($+7.70$), operations, supply chain \& manufacturing ($+7.25$), and energy, utilities \& IoT ($+6.21$) have the largest positive offsets; transportation \& mobility ($-11.44$), environment, climate \& geospatial ($-8.50$), and education \& learning analytics ($-6.22$) have the largest negative offsets. Each cell aggregates only $6$--$8$ tasks, so these results describe broad domain patterns rather than reliable within-domain model rankings.

\subsubsection{Cause versus Consequence Tasks}
\label{app:cause-cons}

Table~\ref{tab:cause-consequence} compares the two task routes, pairing cause and consequence tasks within each world. The minimum detectable difference is $\mathrm{MDE}=2.802\,s_{\Delta}/\sqrt{n}$, where $n$ counts paired worlds. No model's reported difference reaches its MDE; the per-model differences range from $-3.28$ to $+5.03$ points. At this resolution, the table provides no evidence of a systematic route difficulty difference; it does not establish equal difficulty and is not a model ranking.

\begin{table}[t]
  \centering
  \small
  \setlength{\tabcolsep}{4.5pt}
  \caption{Cause versus consequence scores (50 tasks each), following Table~\ref{tab:bench-results}. $\Delta$ = cause $-$ consequence. MDE uses same-world pairs. No model's $\Delta$ reaches its MDE.}
  \label{tab:cause-consequence}
  \resizebox{\columnwidth}{!}{%
  \begin{tabular}{lcccc}
    \toprule
    & \multicolumn{2}{c}{Score ($\times 100$)} & \multicolumn{2}{c}{Paired cause$-$consequence test} \\
    \cmidrule(lr){2-3} \cmidrule(lr){4-5}
    Model & Cause & Consequence & $\Delta$ & MDE \\
    \midrule
    \modellogo{openai}GPT-6-Astra          & $76.32$ & $72.54$ & $+3.78$ & $9.30$ \\
    \modellogo{qwen}Qwen-3.8-Max-0902    & $75.32$ & $71.53$ & $+3.80$ & $8.79$ \\
    \modellogo{openai}GPT-6.1-Sol          & $73.33$ & $73.27$ & $+0.07$ & $8.87$ \\
    \modellogo{zai}GLM 5.3              & $66.05$ & $69.33$ & $-3.28$ & $9.06$ \\
    \modellogo{hunyuan}HY4-Preview          & $68.02$ & $65.08$ & $+2.94$ & $9.81$ \\
    \modellogo{kimi}Kimi-K3              & $68.91$ & $63.88$ & $+5.03$ & $10.35$ \\
    \modellogo{deepseek}DeepSeek-V4.1-Flash  & $64.10$ & $61.80$ & $+2.30$ & $10.25$ \\
    \modellogo{qwen}Qwen-3.8-27B   & $60.40$ & $58.87$ & $+1.53$ & $10.38$ \\
    \modellogo{gemini}Gemini-3.8-Flash     & $55.96$ & $56.38$ & $-0.43$ & $9.52$ \\
    \modellogo{minimax}MiniMax-M3           & $50.56$ & $48.94$ & $+1.62$ & $10.43$ \\
    \bottomrule
  \end{tabular}}%
\end{table}

\subsection{RL Training Implementation Details}
\label{app:rl-training}

\subsubsection{Training Data and Software}
DataWorld-35B-A3B uses 2{,}135 tasks from 1{,}037 distinct worlds, all disjoint from the test worlds. For a fair comparison, we retrain DataMind-35B-A3B on 7{,}323 publicly released Python data-analysis tasks with the same backbone and training configuration, as the original model uses a different backbone and a context window too short for our evaluation.

We train \textbf{DataWorld-35B-A3B} from the base Qwen3.6-35B-A3B checkpoint, without a task-specific supervised fine-tuning stage. The training stack uses slime v0.3.1 (source revision \texttt{a74ae3a0}), Megatron-LM v0.16.0rc0 (\texttt{1dcf0daf}), and SGLang v0.5.15.post1 (\texttt{0b3bb0cb}) for rollout generation. The versions were measured in the training container; rollout and the associated inference services use the same SGLang installation. This paragraph describes the training rollout stack, not the mini-swe-agent evaluation harness used for Table~\ref{tab:base-vs-rl}.

\subsubsection{Resources and Optimization}
Table~\ref{tab:rl-training-settings} summarizes the GPU allocation, parallelism, batch sizes, and optimization settings.

\begin{table}[t]
  \centering
  \small
  \setlength{\tabcolsep}{4pt}
  \caption{RL training resources and settings for the training and rollout workers.}
  \label{tab:rl-training-settings}
  \begin{tabular}{@{}p{0.45\columnwidth}p{0.44\columnwidth}@{}}
    \toprule
    Setting & Value \\
    \midrule
    Training / rollout GPUs & 48 / 64 H20-3e GPUs (6 / 8 eight-GPU nodes) \\
    Training parallelism & TP 2, CP 4, PP 1, EP 8, ETP 1 \\
    Derived data parallelism & DP $=48/(2\times4\times1)=6$ \\
    Rollout engines & 64 one-GPU SGLang engines (TP 1) \\
    Rollout batch size & 24 prompts \\
    Samples per prompt & 8 \\
    Global batch size & 192 sampled responses ($24\times8$) \\
    Micro-batch size & 1 \\
    Rollout steps per update & 1 \\
    Maximum context / generation & 262{,}144 / 16{,}384 tokens \\
    Optimizer and learning rate & Adam, constant $10^{-6}$ \\
    Weight decay / Adam $\beta$ & 0.1 / (0.9, 0.98) \\
    GSPO clip (low / high) & $3.5\times10^{-4}$ / $5\times10^{-4}$ \\
    KL loss / entropy coefficients & 0 / 0 \\
    Rollout temperature & 1.0 \\
    \bottomrule
  \end{tabular}
\end{table}

\subsubsection{Rollout Environment}
Each training rollout executes its agent interactions in a separate, isolated sandbox. The sandbox image includes a Python environment and common data-analysis packages for inspecting and processing the task files; model generation is served separately by the SGLang rollout engines described above. The training uses GSPO with group size eight and rubric-derived rewards scored by the Qwen3.6-35B-A3B base model itself to reduce reward-scoring costs, as described in the main experiments. The settings above describe the training run; the paired evaluation uses a separate, shared mini-swe-agent protocol detailed in Section~\ref{app:rl-eval}.

\subsection{RL Evaluation Protocol}
\label{app:rl-eval}

This section details the evaluation settings behind Table~\ref{tab:base-vs-rl}, with paired comparisons focused on DataWorld-35B-A3B versus its base. We evaluate the models on four benchmarks that probe complementary data-agent abilities: \textbf{JobBench}, which tests source-grounded professional reasoning across a broad range of occupations~\citep{li2026jobbench}; \textbf{DAComp}, which measures open-ended data-science task solving graded by LLM judges; \textbf{DABstep-Hard}, which tests multi-step reasoning over real-world tabular data with strictly verifiable answers; and the held-out split of our own \textbf{DataWorld-Bench}, which stresses open-ended data computation and insight discovery inside fully executable worlds.

\subsubsection{Shared Evaluation and Scoring}
Within each benchmark, all models in Table~\ref{tab:base-vs-rl} are evaluated under the same mini-swe-agent harness~\citep{swebench2025minisweagent}, budget chain, and scoring procedure. DAComp uses Gemini-3.8-Flash as its rubric-based judge; DABstep-Hard uses automatic correctness checks, while DataWorld-Bench applies its own rubrics with the same Gemini-3.8-Flash judge. For the judge-graded DAComp and DataWorld-Bench tasks, scores come from a single judging pass and are scaled by $100$ for display; DataWorld-Bench normalizes net rubric points to $[0,1]$, while DAComp retains its native scoring. DABstep-Hard instead uses binary correctness. Tables~\ref{tab:base-vs-rl} and~\ref{tab:bench-results} use the same DataWorld-Bench judge, rubrics, and single-pass scoring procedure for archived reports, so their DataWorld-Bench scores share the same scoring scale. The tables cover distinct sets of evaluated models; Gemini-3.8-Flash is the shared judge, not an overlapping evaluated model. The comparisons in Table~\ref{tab:base-vs-rl} are internally consistent because each pair shares the same benchmark-specific scoring procedure.

\subsubsection{Paired Statistical Analysis}
Tasks are paired one-to-one by identical task name across the two systems, and $\Delta$ is the per-task difference DataWorld-35B-A3B $-$ base. The paired analysis covers DAComp, DABstep-Hard, and DataWorld-Bench, with $200$, $378$, and $100$ complete pairs, respectively; no task is dropped from these pairings. The paired one-sample test on the per-task differences yields the minimum detectable difference $\mathrm{MDE} = 2.802\,s_{\Delta}/\sqrt{n}$, the same formula used in Section~\ref{app:cause-cons}.

\subsubsection{Benchmark-Specific Details}

\paragraph{JobBench ($n=65$)}~\citep{li2026jobbench} evaluates agents on professional work tasks that domain experts most want delegated, spanning 35 occupations from office support to management and life science. Each task packages heterogeneous reference files, and responses are graded by fact-anchored chains of binary rubric criteria. We evaluate the 65-task main set, which mixes real public records with synthesized documents and poses more reasoning challenges than the easy set. Its scope extends beyond data analysis, making JobBench the least task-aligned of the four benchmarks and a probe of transfer to broader file-based professional reasoning; aggregate gains do not establish improvement within every occupation.

\paragraph{DAComp ($n=200$)}~\citep{lei2026dacomp} mirrors enterprise data-intelligence workflows with 210 tasks in two categories: \emph{data engineering} tasks that require repository-level engineering on industrial schemas---designing multi-stage SQL pipelines from scratch or evolving existing systems under changing requirements---and \emph{data analysis} tasks that pose open-ended business problems demanding strategic planning, iterative exploratory coding, and the synthesis of actionable recommendations. Engineering tasks are scored by execution-based multi-metric evaluation, while open-ended tasks are graded by an experimentally validated LLM judge guided by hierarchical rubrics. We evaluate only the open-ended data-analysis subset ($n=200$), using Gemini-3.8-Flash to judge both systems against the same rubrics, so both systems receive fine-grained, rubric-based scores rather than binary correctness labels.

\paragraph{DABstep-Hard ($n=378$)}~\citep{egg2025dabstep} is the hardest difficulty split of DABstep, a benchmark of over 450 real-world data-analysis tasks derived from a financial analytics platform, in which the agent must combine code-based data processing with contextual reasoning over heterogeneous documentation. Tasks use a factoid answer format with automatic correctness checks, and difficulty is explicitly stratified---even the strongest agents report accuracy below $15\%$ on the hard split. Each task is therefore graded pass/fail: the reward takes only the values $0$ and $1$, so the reported mean is a pass rate.

\paragraph{DataWorld-Bench ($n=100$).} The held-out test split of our own benchmark ($50$ worlds, one cause and one consequence task each). Beyond producing a final answer, the agent must navigate the full world folder, compute intermediate variables through iterative code execution, and derive insights from the results, and its report is graded by rubric-based LLM judges with both positive and negative items.

\paragraph{MMLU-Pro.} MMLU-Pro~\citep{wang2024mmlupro} evaluates broad knowledge and reasoning through multiple-choice questions across diverse academic and professional subjects. Compared with MMLU, it emphasizes more challenging, reasoning-oriented questions and expands the answer choices. We use it as a general-capability retention check and report accuracy (\%) for DataWorld-35B-A3B and its base in Table~\ref{tab:general-capabilities}.

\paragraph{LiveCodeBench-Hard.} LiveCodeBench~\citep{jain2024livecodebench} evaluates code generation on programming problems drawn from coding competitions, using executable test cases to assess solution correctness. We evaluate its hard subset as a probe of general algorithmic coding ability beyond the data-analysis tasks used for training. Table~\ref{tab:general-capabilities} reports accuracy (\%); this check is distinct from rubric-based scoring of open-ended analytical reports.

\paragraph{SimpleQA-Verified.} SimpleQA Verified~\citep{haas2025simpleqaverified} evaluates short-form factual question answering using a curated subset of OpenAI's SimpleQA, with de-duplication, topic balancing, source reconciliation, and an improved autorater prompt. We use it as a complementary factuality check and report F1 in Table~\ref{tab:general-capabilities}. It probes factual answering outside the synthesized data environments, rather than directly measuring unsupported claims about those worlds; a higher score alone does not establish a general reduction in hallucinations.

\end{document}